\pdfoutput=1

\documentclass[11pt]{article}

\usepackage[]{acl}

\usepackage{times}
\usepackage{latexsym}
\usepackage{tabularx,booktabs}
\usepackage{rotating}
\usepackage{graphicx}

\usepackage{caption}
\usepackage{subcaption}

\usepackage[T1]{fontenc}

\usepackage[utf8]{inputenc}

\usepackage{microtype}

\usepackage{graphicx}
\usepackage{booktabs}
\usepackage{longtable}
\usepackage{tabu}
\usepackage{listings}
\usepackage{hyperref}
\usepackage{tcolorbox}

\title{An Evaluation of Persian-English Machine Translation Datasets with Transformers}

\author{
  Amir Sartipi \\ University of Isfahan\\ \texttt{amirsartipi.msc@eng.ui.ac.ir} \\ \And
  Meghdad Dehghan \\ University of Isfahan\\ \texttt{meghdadd78@gmail.com} \\ \AND
  Afsaneh Fatemi \\ University of Isfahan \\ \texttt{a\_fatemi@eng.ui.ac.ir}
}

\begin{document}
\maketitle
\begin{abstract}
Nowadays, many researchers are focusing their attention on the subject of machine translation (MT). However, Persian machine translation has remained unexplored despite a vast amount of research being conducted in languages with high resources, such as English. Moreover, while a substantial amount of research has been undertaken in statistical machine translation for some datasets in Persian, there is currently no standard baseline for transformer-based text2text models on each corpus. This study collected and analysed the most popular and valuable parallel corpora, which were used for Persian-English translation. Furthermore, we fine-tuned and evaluated two state-of-the-art attention-based seq2seq models on each dataset separately (48 results). We hope this paper will assist researchers in comparing their Persian to English and vice versa machine translation results to a standard baseline.
\end{abstract}

\begin{figure}[ht]
        \centering
        \includegraphics[width= 0.95 \linewidth]{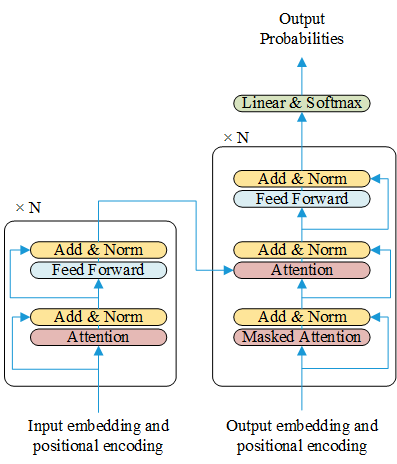}
        \caption{Transformer model architecture}
        \label{fig:ch1-transformer}
\end{figure}

\section{Introduction}

The primary purpose of machine translation is to translate texts from one language to another. Previously a statistical language model used to be considered as the frontier of this task \citep{brown-etal-1993-mathematics, koehn2009statistical, lopez2008statistical}. However, because of the vast amount of data currently available, neural machine translation \citep{bahdanau2014neural, kalchbrenner2013recurrent, wu2016google, cho2014properties} is now surpassing statistical approaches. Then, a new simple network architecture based solely on attention was proposed by \citet{vaswani2017attention} as an alternative to the dominant sequence transduction models based on recurrent and convolutional neural networks. The encoder part of transformer architecture has been widely used in \citet{devlin2018bert} and \citet{liu2019roberta} which pre-trained on large amount of unlabeled text. \citet{raffel2020exploring} examined the landscape of transfer learning strategies for NLP resulting in the emergence of transfer learning as a potent technique in NLP. It presents a system that transforms all language tasks into text-to-text format which is called T5. The mT5 is a multilingual variant of the T5 model that has been pre-trained with a new Common Crawl-based dataset that contains 101 languages \citep{xue-etal-2021-mt5}. In order to combat overfitting while training on thousands of tasks, \citet{costa2022no} proposed multiple architectural and training improvements. They used a human-translated benchmark, Flores-200, to evaluate the performance of over 40,000 different translation directions. Compared to the previous state-of-the-art seq2seq models, their model achieved a 44\% improvement in BLEU Score. Both of these two models (google T5 and meta NLLB) utilize the transformer architecture with some changes and improvements in the encoder or the decoder part. 
The transformer architecture is shown in Figure \ref{fig:ch1-transformer}. 

The purpose of this paper can be summarized as follows:
\begin{enumerate}
    \item We review statistical and neural machine translation systems and related datasets.
    \item We release all experiments results, including last model checkpoint, best model checkpoint, model prediction, history of training and development phase, and execution times are publicly available in Hugging Face\footnote{\url{https://huggingface.co/}} and also codes are available in the GitHub\footnote{\url{https://github.com/}} repository.
    \item We establish baselines for the Persian-English machine translation task to compare by future research.
    \item We investigate the influence of the number of instances on the BLEU score.
\end{enumerate}

The rest of the article is structured in the following manner. In section \ref{sec:rw} we summarize prior approaches to translating Persian-English machine translation. Section \ref{sec:data} explains the most popular corpora which are used for experiments. In addition to that, we also provide a detailed analysis of their statistics in this section. An extensive set of experiments with language models are provided in section \ref{sec:expr} for each dataset, and they are conducted in both directions. In section \ref{sec:disc}, the challenges of the study are argued and an analysis of models' predictions is provided. Finally, in Section \ref{sec:conc}, the conclusions of the study are presented.

\begin{figure*}[htp]
        \centering
        \includegraphics[width= 0.95 \linewidth]{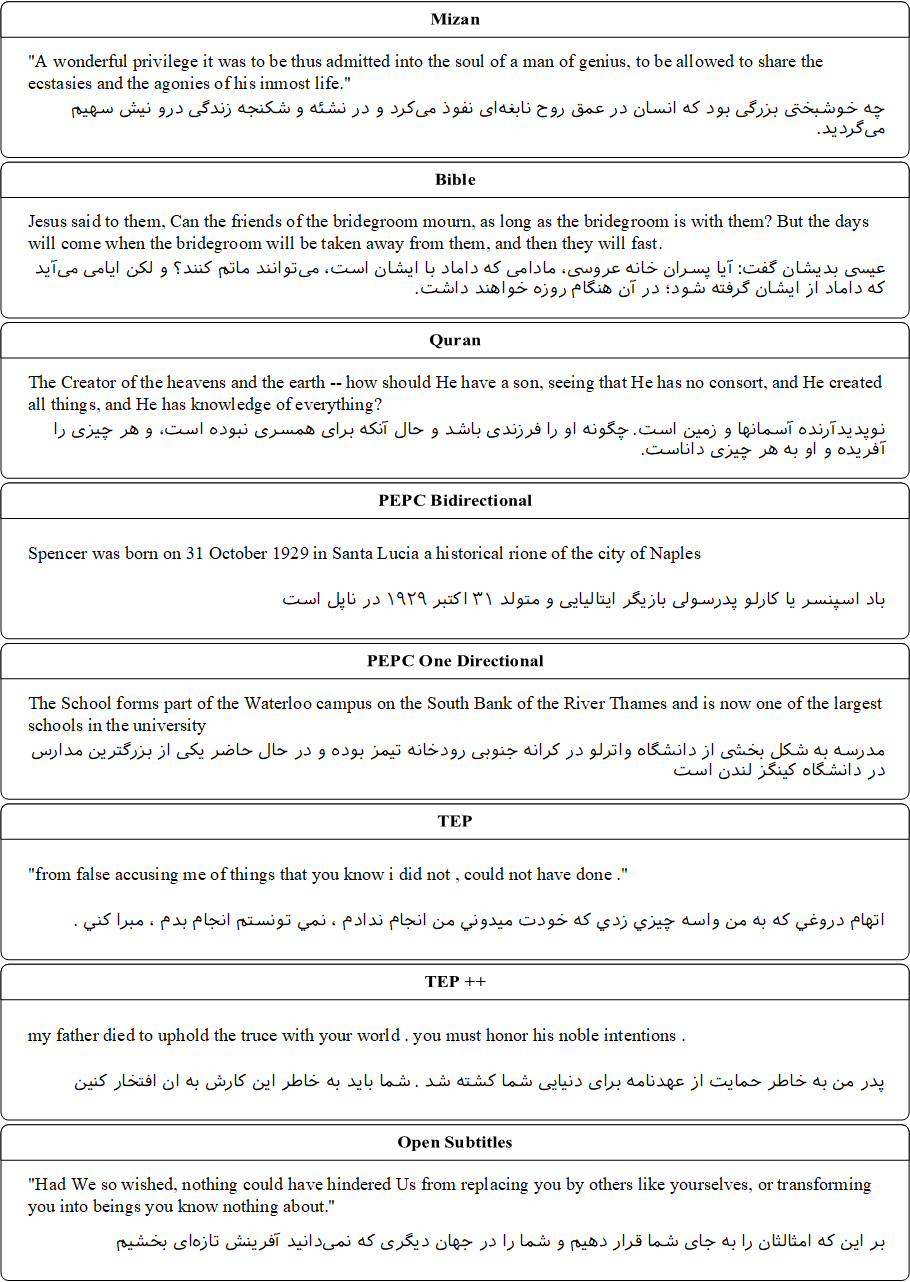}
        \caption{Examples of English (top) and Persian (bottom) side instances for each dataset}
        \label{fig:ch3-examples}
\end{figure*}

\begin{table*}[]
\centering
\resizebox{\textwidth}{!}{%
\begin{tabular}{c|cccccc|cccccc|}
\cline{2-13}
                                         & \multicolumn{6}{c|}{\textbf{Persian}}                                                                                                                                                                & \multicolumn{6}{c|}{\textbf{English}}                                                                                                                                                                \\ \cline{2-13} 
                                         & \multicolumn{1}{c|}{\textbf{avg}} & \multicolumn{1}{c|}{\textbf{min}} & \multicolumn{1}{c|}{\textbf{max}} & \multicolumn{1}{c|}{\textbf{92\%}} & \multicolumn{1}{c|}{\textbf{all}} & \textbf{unique} & \multicolumn{1}{c|}{\textbf{avg}} & \multicolumn{1}{c|}{\textbf{min}} & \multicolumn{1}{c|}{\textbf{max}} & \multicolumn{1}{c|}{\textbf{92\%}} & \multicolumn{1}{c|}{\textbf{all}} & \textbf{unique} \\ \hline
\multicolumn{1}{|c|}{\textbf{Mizan}}     & \multicolumn{1}{c|}{13}           & \multicolumn{1}{c|}{1}            & \multicolumn{1}{c|}{232}          & \multicolumn{1}{c|}{26}            & \multicolumn{1}{c|}{13,464,236}   & 131,751         & \multicolumn{1}{c|}{13}           & \multicolumn{1}{c|}{0}            & \multicolumn{1}{c|}{226}          & \multicolumn{1}{c|}{26}            & \multicolumn{1}{c|}{13,360,397}   & 259,182         \\ \hline
\multicolumn{1}{|c|}{\textbf{Bible}}     & \multicolumn{1}{c|}{28}           & \multicolumn{1}{c|}{3}            & \multicolumn{1}{c|}{124}          & \multicolumn{1}{c|}{48}            & \multicolumn{1}{c|}{1,796,084}    & 18,166          & \multicolumn{1}{c|}{23}           & \multicolumn{1}{c|}{2}            & \multicolumn{1}{c|}{100}          & \multicolumn{1}{c|}{38}            & \multicolumn{1}{c|}{1,428,716}    & 40,202          \\ \hline
\multicolumn{1}{|c|}{\textbf{Quran}}     & \multicolumn{1}{c|}{29}           & \multicolumn{1}{c|}{1}            & \multicolumn{1}{c|}{373}          & \multicolumn{1}{c|}{61}            & \multicolumn{1}{c|}{30,235,077}   & 28,380          & \multicolumn{1}{c|}{33}           & \multicolumn{1}{c|}{1}            & \multicolumn{1}{c|}{772}          & \multicolumn{1}{c|}{74}            & \multicolumn{1}{c|}{34,227,828}   & 92,976          \\ \hline
\multicolumn{1}{|c|}{\textbf{PEPC Bidirectional}}   & \multicolumn{1}{c|}{20}           & \multicolumn{1}{c|}{7}            & \multicolumn{1}{c|}{178}          & \multicolumn{1}{c|}{35}            & \multicolumn{1}{c|}{4,163,011}    & 169,637         & \multicolumn{1}{c|}{21}           & \multicolumn{1}{c|}{7}            & \multicolumn{1}{c|}{153}          & \multicolumn{1}{c|}{36}            & \multicolumn{1}{c|}{4,354,619}    & 142,792         \\ \hline
\multicolumn{1}{|c|}{\textbf{PEPC One Directional}}  & \multicolumn{1}{c|}{22}           & \multicolumn{1}{c|}{7}            & \multicolumn{1}{c|}{178}          & \multicolumn{1}{c|}{37}            & \multicolumn{1}{c|}{3,539,183}    & 158,707         & \multicolumn{1}{c|}{21}           & \multicolumn{1}{c|}{7}            & \multicolumn{1}{c|}{153}          & \multicolumn{1}{c|}{36}            & \multicolumn{1}{c|}{3,359,635}    & 138,489         \\ \hline
\multicolumn{1}{|c|}{\textbf{TEP}}       & \multicolumn{1}{c|}{8}            & \multicolumn{1}{c|}{1}            & \multicolumn{1}{c|}{37}           & \multicolumn{1}{c|}{14}            & \multicolumn{1}{c|}{716,113}      & 22,710          & \multicolumn{1}{c|}{7}            & \multicolumn{1}{c|}{1}            & \multicolumn{1}{c|}{33}           & \multicolumn{1}{c|}{14}            & \multicolumn{1}{c|}{684,242}      & 36,634          \\ \hline
\multicolumn{1}{|c|}{\textbf{TEP ++}}     & \multicolumn{1}{c|}{7}            & \multicolumn{1}{c|}{1}            & \multicolumn{1}{c|}{34}           & \multicolumn{1}{c|}{13}            & \multicolumn{1}{c|}{4,445,543}    & 92,037          & \multicolumn{1}{c|}{8}            & \multicolumn{1}{c|}{0}            & \multicolumn{1}{c|}{32}           & \multicolumn{1}{c|}{14}            & \multicolumn{1}{c|}{4,720,821}    & 57,753          \\ \hline
\multicolumn{1}{|c|}{\textbf{OPUS-100}}      & \multicolumn{1}{c|}{10}           & \multicolumn{1}{c|}{1}            & \multicolumn{1}{c|}{1,487}        & \multicolumn{1}{c|}{21}            & \multicolumn{1}{c|}{10,284,744}   & 155,874         & \multicolumn{1}{c|}{9}            & \multicolumn{1}{c|}{1}            & \multicolumn{1}{c|}{839}          & \multicolumn{1}{c|}{20}            & \multicolumn{1}{c|}{9,524,220}    & 342,979         \\ \hline
\end{tabular}%
}
\caption{General statistics for datasets}
\label{general-information}
\end{table*}

\section{Related Work}
\label{sec:rw}

As far as previous research is concerned, there have been several studies conducted for English to Arabic \citep{turjuman}, French \citep{TIAN20221438, french2, french3}, and Russian \citep{ru1, ru2}, which focus on transformers as a basic architecture and represent results. However, there are some Persian-English datasets without any results on language models.

In this section we will investigate previous works on Persian-English machine translation. First we consider two statistical and neural approaches to machine translation and introduce recent works on these domains. Then we review the attempts in which parallel corpus for Persian-English machine translation was introduced.

\begin{table}[]
\centering
\resizebox{\columnwidth}{!}{%
\begin{tabular}{c|cccc}
                   & \textbf{train} & \textbf{dev} & \textbf{test} & \textbf{all} \\
                   \hline
\textbf{Mizan}     & 1,006,430      & 5,000        & 10,166        & 1,021,596    \\
\textbf{Bible}     & 51,329         & 5,000        & 5,704         & 62,033       \\
\textbf{Quran}     & 1,013,756      & 5,000        & 10,240        & 1,028,996    \\
\textbf{PEPC Bidirectional}   & 175,442        & 5,000        & 19,494        & 199,936      \\
\textbf{PEPC One Directional}  & 138,005        & 5,000        & 15,334        & 158,339      \\
\textbf{TEP}       & 72,748         & 5,000        & 8,084         & 85,832       \\
\textbf{Tep ++}     & 515,925        & 5,000        & 57,326        & 578,251      \\
\textbf{OPUS-100}      & 1,000,000      & 2,000        & 2,000         & 1,004,000    \\
\end{tabular}%
}
\caption{The number of instances in train\textbackslash dev\textbackslash test }
\label{instances-tdt}
\end{table}

\paragraph{Baselines on SMT systems.}
Results for a Persian-English SMT system were first obtained in the PersianSMT  \citep{pilevar2010persiansmt}. They used a phrased-based SMT system and obtained results on the movie subtitle domain as their parallel corpus's main resource. In addition, \citet{bakhshaei2010study} obtained results for phrase-based Persian-English SMT system. Different values of the SMT system parameters were tested, and the results for each parameter value were compared. \citet{mohaghegh2010performance} and \citet{mohaghegh2010improved} achieve results for an SMT system for different sizes of language model corpora. They concluded that training SMT systems with larger corpora led in better results. \citet{mohaghegh2011improving} created a combined parallel corpus called NSPEC and obtained better results for their SMT system than their previous work. 

\citet{pilevar2011using} created a RBSMT system followed by statistical editing and obtained results for their system. Their new approach outperformed the existing RBSMT systems, yet SMT systems were still more effective than their approach. \citet{mohaghegh2012advancements} compared two hierarchical (the Joshua) and classical (the Moses) SMT systems. They obtained results for both directions; however,  using the hierarchical system only in the English-to-Persian translation direction produced better results. 

\citet{afec} created a new corpus whose obtained results for SMT systems outperformed the previous ones. \citet{mansouri2012state} compared several SMT systems and also used a maxent classifier to refine the existing state-of-the-art SMT system.  \citet{rasooli2013orthographic} showed that segmenting Persian verbs is effective and improves the BLEU score. \citet{tep++} improved exiting TEP corpus and created TEP++. They also gained results on their new corpus and compared them to other corpora like TEP and Mizan. The findings of their study surpassed previous results on both TEP and Mizan corpora. In their study, \citet{mizan} calculated BLEU score for the SMT system on their represented corpus (Mizan). They achieved results for both in-domain and out-of-domain test sets.
\paragraph{Baselines on NMT systems.}
Several attempts have been made to propose baselines on Persian-English machine translation using neural machine translation systems. \citet{bastan2017neural} conducted a study on two tasks of translation and transliteration using a neural machine translation (NMT) system. They used RNNs in the NMT architecture for different numbers of layers. Additionally, they enhanced the results by changing the cost function and preprocessing the Persian corpus. Compared with existing NMT systems, \citet{zaremoodi2018adaptive} and \citet{zaremoodi2018neural} demonstrated that a multi-task-learning approach improves machine translation results for low-resource languages like Persian. 
PasriNLU used a neural language model for the first time to do machine translation between Persian and English \citep{parsinlu}. They fine-tuned four variations of the Google mT5 text2text model on a part of a benchmark that they created. The training dataset used in the fine-tuning process was integrated from four corpora for generalisation purposes.

\section{Datasets}
\label{sec:data}

The vast majority of research and benchmarks on the machine translation task have been done on the WMT dataset \citep{wmt}. 
Also there are datasets like OPUS-100 \citep{opus-100} and OpenSubtitles \citep{opensub} which contain 60 and 100 languages respectively and are used in the machine translation task for other languages.

For the Persian-English language pair, we have collected nine datasets to be fine-tuned with neural seq2seq and to gain results for each of them. Moreover, ParsiNLU is a set of language understanding tasks, including machine translation, for the Persian language \citep{parsinlu}. In the machine translation part of their work, they created a large parallel corpus integrated from several corpora. The training dataset includes four domains: the questions from their question paraphrasing task, the Mizan corpus, the TEP corpus and the Global Voice corpus. The training dataset contains almost 1.6M entries. The evaluation set consists of Quran, Mizan, Bible and QPP datasets and contains about 47k sentences. 

Each collected dataset is introduced and their main attributes are investigated as follows. 

\begin{figure}[h]
    \centering

      \begin{subfigure}[b]{1\columnwidth}
        \includegraphics[width=\linewidth]{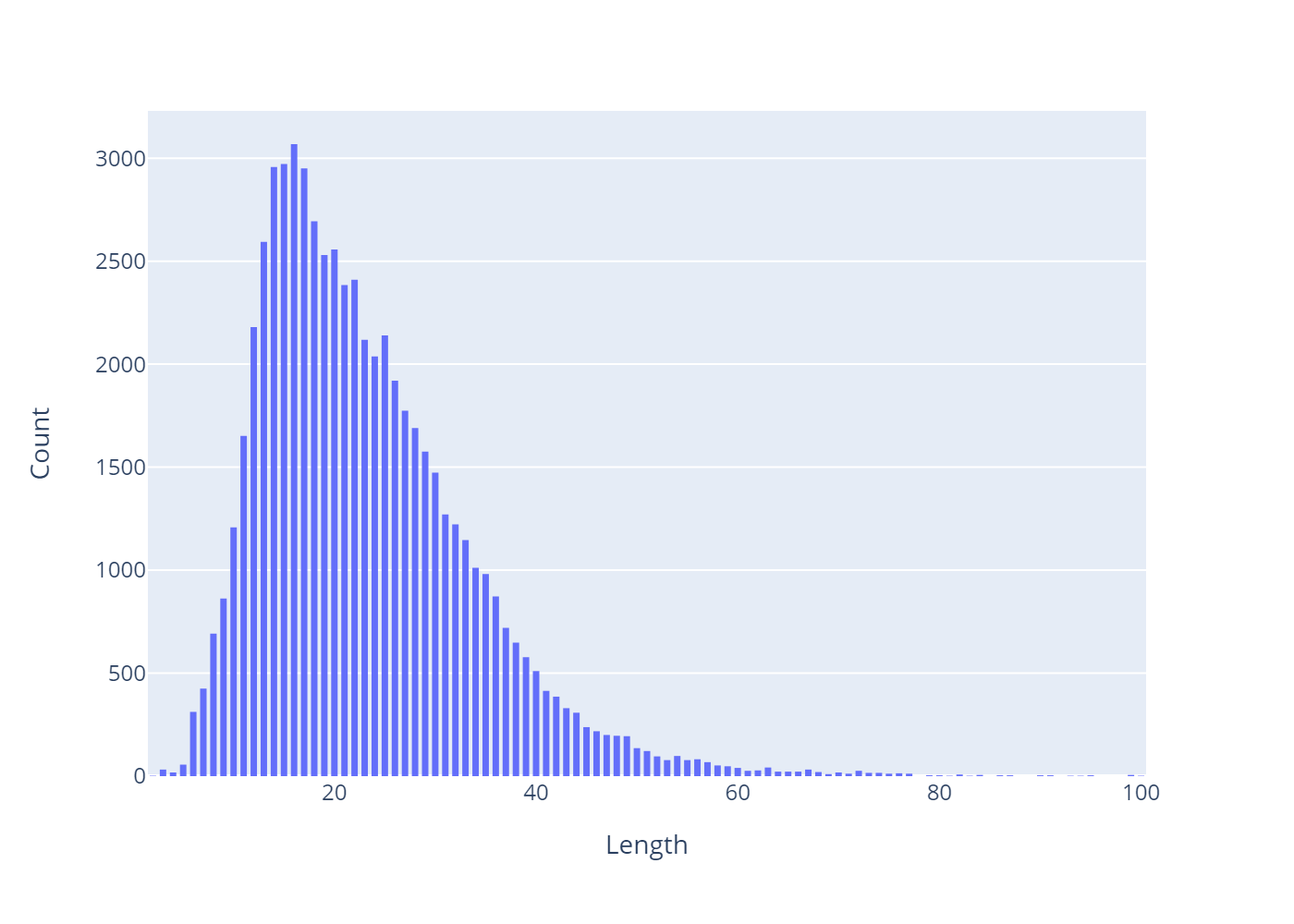}
        \caption{English side}
      \end{subfigure}
      \begin{subfigure}[b]{1\columnwidth}
        \includegraphics[width=\linewidth]{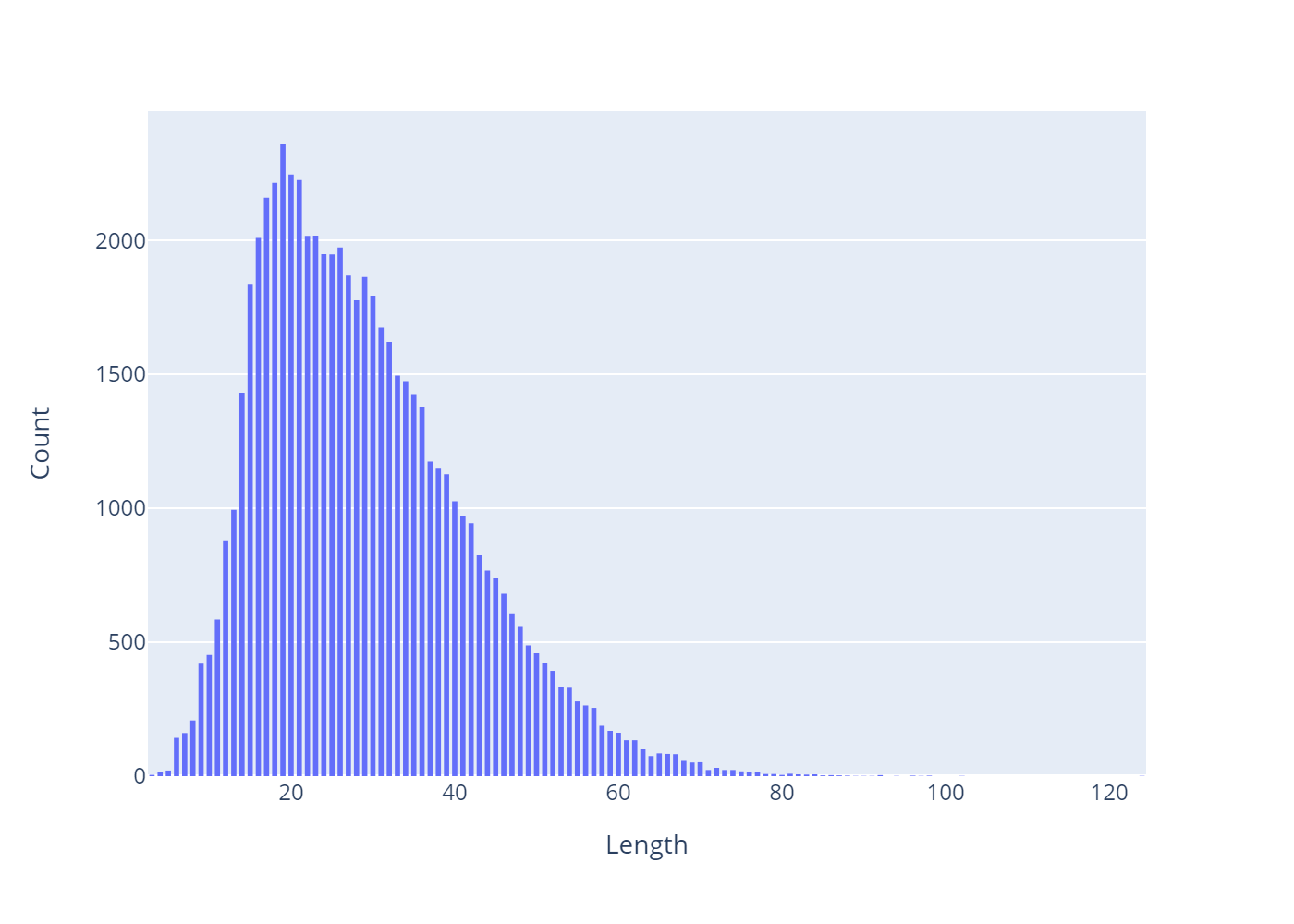}
        \caption{Persian side}
      \end{subfigure}

    \caption{Token distribution per sentences for Bible}
    \label{fig:word_distru}
\end{figure}


\paragraph{Quran.}
Quran is primarily an Arabic book which has been translated into many languages. \citet{tiedemann2012parallel} proposed the Tanzil dataset from the Tanzil project as a part of the OPUS project. This dataset contains 42 languages. The Persian-English language pair of this dataset contains almost 1M sentence pairs and 57.02M words. 

\paragraph{Bible.}
Bible is another religious book which has been translated into many languages. As a part of the OPUS project, the Bible dataset was released in 100 languages \citep{tiedemann2012parallel}. The Persian-English language pair of this dataset contains almost 62,000 sentence pairs and 2.89M words.

\paragraph{TEP.}
TEP (Tehran English Persian) is another parallel corpus made from movie subtitles. Almost 21000 subtitle files were collected from Open-subtitles, and only 1200 subtitle file pairs remained after removing duplicate files. The final dataset contains over 550,000 lines of text \citep{pilevar2011tep}.

\paragraph{TEP++.}
A refined version of the TEP corpus named TEP++ was introduced by \citet{tep++}. They reported that the TEP corpus was noisy, and they tried to fix this problem in the new corpus. They also obtained better results for an SMT system by using the TEP++ corpus. This corpus has near 570,000 aligned sentences and near 5M tokens for both Persian and English languages.

\paragraph{OPUS-100.}
OPUS-100 is a concatenation of movie subtitles, GNOME documentation, and Bible datasets that contains 100 languages and 99 language pairs, all of which use English as a source or target language \citep{opus-100}.

\paragraph{PEPC.}
PEPC is another parallel corpus for Persian-English language pairs obtained from Wikipedia documents \citep{pepc}. They used bidirectional and one-directional methods to extract documents from Wikipedia, so they proposed two versions of datasets based on the extraction method. The bidirectional PEPC dataset contains near 200,000 sentence pairs, and the one-directional PEPC dataset contains near 160,000 sentence pairs.

\paragraph{Mizan.}
Mizan was the largest Persian corpus at the time it was released. It was created from literature masterpieces. It contains more than one million sentence pairs and over 23M words for both Persian and English \citep{mizan}.

We randomly selected an instance from each corpus which is illustrated in Figure \ref{fig:ch3-examples}. It appears that the OPUS-100 dataset places capitalized "We" and "Us," in the middle of a sentence, a dictation mistake in the Persian subtitle, and the word-by-word translation and its meaning is not perfectly aligned. Some sentences are enclosed in quotation marks or start with small letters in English. These features of datasets could affect the evaluation results.

We used SPARK NLP \citep{sparknlp} to provide general statistical information about datasets. As a result of this information, parameters such as sequence lengths can be selected more precisely. The max column in table \ref{general-information} indicates the maximum number of tokens that are allowed in a sentence. Because each dataset contains a few long sequences that can be chosen as outliers and could be simply truncated by a more precise length, this number may not be a good choice. Therefore, for each dataset, we calculated a number which covers 92 percent of datasets. In other words, 92\% of sentences have a less or an equal number of tokens. In terms of tokens per sentence, this number is much lower than the maximum. In addition, the table contains both the average and the minimum number of tokens per dataset, as well as the total number of tokens and the total number of unique tokens for both Persian and English corpora.

\section{Experiments}
\label{sec:expr}

\begin{table*}[]
\centering
\resizebox{\textwidth}{!}{%
\begin{tabular}{c|ccc|ccc|}
\cline{2-7}
                                                    & \multicolumn{3}{c|}{\textbf{EN-FA}}                                                                        & \multicolumn{3}{c|}{\textbf{FA-EN}}                                                                        \\ \cline{2-7} 
                                                    & \multicolumn{1}{c|}{\textbf{mt5-small}} & \multicolumn{1}{c|}{\textbf{mt5-base}} & \textbf{nllb-distilled} & \multicolumn{1}{c|}{\textbf{mt5-small}} & \multicolumn{1}{c|}{\textbf{mt5-base}} & \textbf{nllb-distilled} \\ \hline
\multicolumn{1}{|c|}{\textbf{Mizan}}                & \multicolumn{1}{c|}{12.22}              & \multicolumn{1}{c|}{12.69}             & 15.00                   & \multicolumn{1}{c|}{16.29}              & \multicolumn{1}{c|}{16.70}             & 18.05                   \\ \hline
\multicolumn{1}{|c|}{\textbf{Bible}}                & \multicolumn{1}{c|}{13.93}              & \multicolumn{1}{c|}{22.06}             & 69.78                   & \multicolumn{1}{c|}{16.28}              & \multicolumn{1}{c|}{18.83}             & 49.93                   \\ \hline
\multicolumn{1}{|c|}{\textbf{Quran}}                & \multicolumn{1}{c|}{4.79}               & \multicolumn{1}{c|}{4.97}              & 18.10                   & \multicolumn{1}{c|}{10.39}              & \multicolumn{1}{c|}{10.04}             & 27.65                   \\ \hline
\multicolumn{1}{|c|}{\textbf{PEPC Bidirectional}}   & \multicolumn{1}{c|}{7.10}               & \multicolumn{1}{c|}{7.21}              & 13.13                   & \multicolumn{1}{c|}{10.28}              & \multicolumn{1}{c|}{10.22}             & 17.01                   \\ \hline
\multicolumn{1}{|c|}{\textbf{PEPC One Directional}} & \multicolumn{1}{c|}{5.37}               & \multicolumn{1}{c|}{5.71}              & 13.20                   & \multicolumn{1}{c|}{8.82}               & \multicolumn{1}{c|}{9.85}              & 16.84                   \\ \hline
\multicolumn{1}{|c|}{\textbf{TEP}}                  & \multicolumn{1}{c|}{11.70}              & \multicolumn{1}{c|}{14.11}             & 16.06                   & \multicolumn{1}{c|}{13.63}              & \multicolumn{1}{c|}{23.64}             & 26.74                   \\ \hline
\multicolumn{1}{|c|}{\textbf{TEP ++}}               & \multicolumn{1}{c|}{21.02}              & \multicolumn{1}{c|}{23.09}             & 26.44                   & \multicolumn{1}{c|}{30.14}              & \multicolumn{1}{c|}{31.63}             & 35.98                   \\ \hline
\multicolumn{1}{|c|}{\textbf{OPUS-100}}             & \multicolumn{1}{c|}{10.81}              & \multicolumn{1}{c|}{10.46}             & 11.62                   & \multicolumn{1}{c|}{20.66}              & \multicolumn{1}{c|}{20.91}             & 24.16                   \\ \hline
\end{tabular}%
}

\caption{Evaluation of English to Persian (EN-FA) and Persian to English (FA-EN) on the language
models}
\label{final-results}
\end{table*}

In order to build our network, we used PyTorch \citep{pytorch} and Transformers library from Hugging Face \citep{transformers} as implementation tools.

\paragraph{Datasets' splits.} Table \ref{instances-tdt} provides information about the total number of instances and train/dev/test splits of each dataset.
We used predefined data splits for OPUS-100 dataset. For others we manually split the whole datasets in train/dev/test splits. First we shuffled whole instances of each dataset to randomize their order. Then, for the datasets with more than one million instances, we chose 1\% of whole instances for the test split, 5,000 instances for the dev split and other instances as train split.  

\paragraph{Hyper-Parameters:} \citet{parsinlu} use 1e-3 learning-rate (lr) for fine-tuning phase. The same lr and fine-tuned models for 7 epochs with ADAMW optimizer was used in this study \citep{adamw}. In order to select sequence length during the training phase, we considered what sequence length includes 92\% of our dataset. Besides the number of sentences versus the number of tokens in each sentence were drawn which allowed us to select reasonable sequence length.
Figure \ref{fig:word_distru} shows an example of this illustration for Bible dataset.

\begin{figure}[h]
  \begin{subfigure}[b]{1\columnwidth}
    \includegraphics[width=\linewidth]{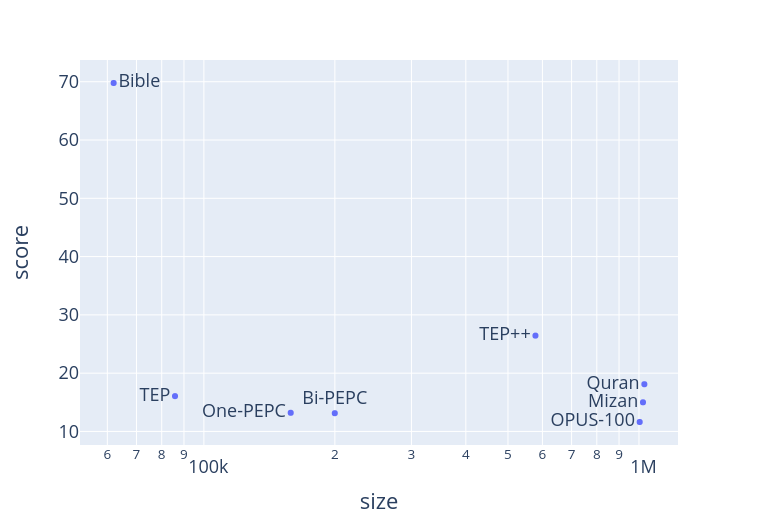}
    \caption{English to Persian direction}
    \label{fig:1}
  \end{subfigure}
  \begin{subfigure}[b]{1\columnwidth}
    \includegraphics[width=\linewidth]{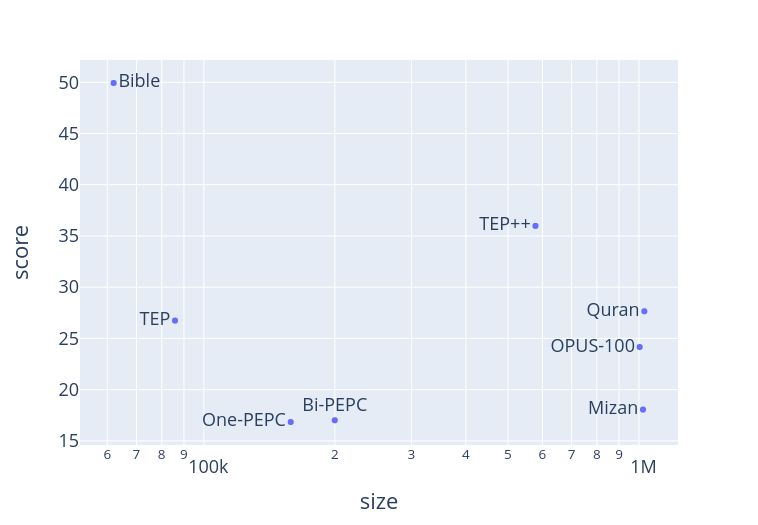}
    \caption{Persian to English direction}
    \label{fig:2}
  \end{subfigure}
   \label{fig:bl-vs-dataset-size}
    \caption{The highest values of BLUE scores according to the datasets' size}
\end{figure}

\paragraph{Models} One of the seq2seq models we used is mT5 which has embedding for Persian language. The other text2text model is NLLB which beats previous cutting-edge models. Because of a huge number of parameters and  the amount of computation power needed for such models, we just fine-tuned datasets on the 2 Google mT5 variants \{mT5 small, mT5 base\} and one Facebook NLLB models: \{distilled NLLB\}. Below we summarize the main attributes of these models
\begin{itemize}
	\item \textbf{Google mT5:} Google T5 model is a text-to-text transformer-based language model. It means that both input and output of this model are text. This model can be used for different tasks such as question answering, machine translation, and text classification. The mT5 version of this model is pre-trained on multi-lingual mC4 data which contains 101 languages including Persian. The mT5-small version of this model is the smallest version with only 300 million parameters. The mT5-base is the second smallest model with 580 million parameters. The largest version of this model has about 13 billion parameters. 
	
	\item \textbf{Meta NLLB:} The NLLB model which is the state-of-the-art text2text model of the time was proposed with the aim of improving the machine translation performance of low-resource languages. It supports embeddings for almost 200 languages. This model also uses a transformer-based architecture and has two types: Dense and MeE. The Dense type is the one that activates all model parameters for each input sequence while the MoE model is the one which activates only a subset of parameters for each input. The NLLB model has 5 variants regrading the size of the parameters. The smallest model has only 600 million parameters and is a Dense model while the largest model which is a MoE model has about 54.5 billion parameters.
\end{itemize}

\paragraph{Evaluation metric:}

The BLEU score \citep{bleu} is the most common metric which has been used for evaluating machine translation results for many years. This metric uses combined N-gram precision for different N-gram sizes and a sentence brevity penalty. Due to the variety of configurations for choosing BLEU score parameters, the results of different baselines by researchers are not much reliable to be compared. For example in many researches, the size of maximum N-gram and the tokenization method is not reported. The sacreBLEU metric was proposed by \citet{sacrebleu} to tackle some of these problems and establish a standard metric to be comparable in different researches.

\paragraph{Training process:}
We considered one direction for each experiment since a model can be fine-tuned simultaneously in Persian and English. The model was evaluated at the end of each epoch during the training phase. The optimum models were selected based on the value of the evaluation metric on the development dataset. It is important to pre-process data before training the models, but we did not do that since we wanted to establish baselines for these datasets. MT systems can be improved by applying data-cleaning approaches to a dataset.

\begin{figure*}[htp]
    \centering
     \begin{subfigure}[b]{1\textwidth}
         \centering
         \includegraphics[width=0.45\linewidth]{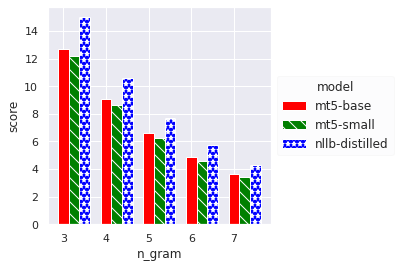}
         \includegraphics[width=0.45\linewidth]{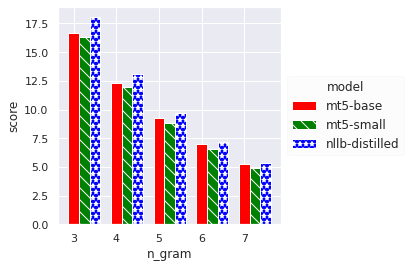}
         \caption{Mizan}
         \label{fig:mizan-ngram}
     \end{subfigure}
     \begin{subfigure}[b]{1\textwidth}
         \centering
         \includegraphics[width=0.45\linewidth]{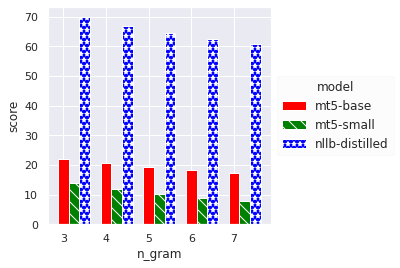}
         \includegraphics[width=0.45\linewidth]{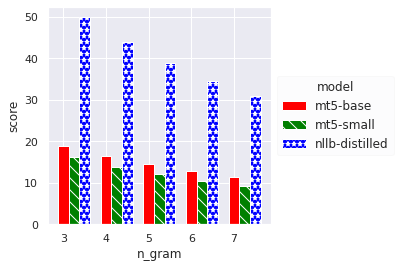}
         \caption{Bible}
         \label{fig:bible-ngram}
     \end{subfigure}

     \begin{subfigure}[b]{1\textwidth}
         \centering
         \includegraphics[width=0.45\linewidth]{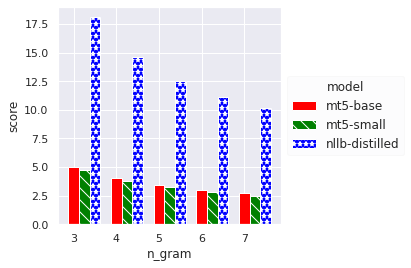}
         \includegraphics[width=0.45\linewidth]{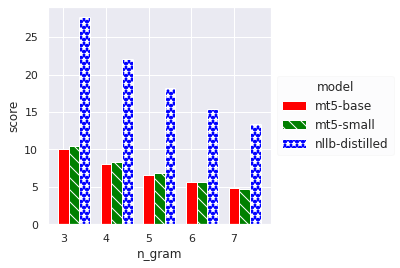}
         \caption{Quran}
         \label{fig:quran-ngram}
     \end{subfigure}

     \begin{subfigure}[b]{1\textwidth}
         \centering
         \includegraphics[width=0.45\linewidth]{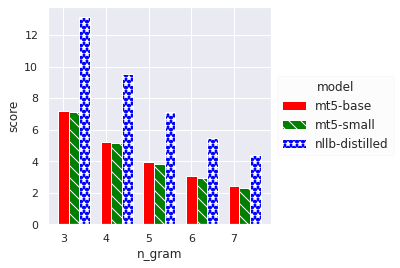}
         \includegraphics[width=0.45\linewidth]{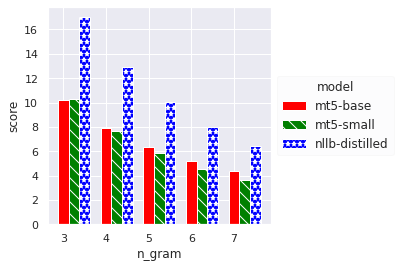}
         \caption{PEPC Bidirectional}
         \label{fig:pepc-bi-ngram}
     \end{subfigure}
    
    \caption{BLEU Score results for different ngrams separated by translation direction (left side English to Persian and right side Persian to English) and model \textbf{First part}}

\end{figure*}

\begin{figure*}[htp]
    \centering
    \ContinuedFloat
     \begin{subfigure}[b]{1\textwidth}
         \centering
         \includegraphics[width=0.45\linewidth]{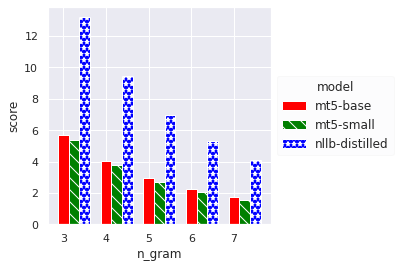}
         \includegraphics[width=0.45\linewidth]{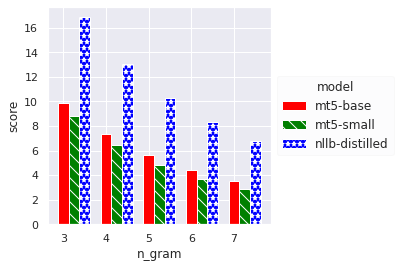}
         \caption{PEPC One Directional}
         \label{fig:pepc-one-ngram}
     \end{subfigure}
     \begin{subfigure}[b]{1\textwidth}
         \centering
         \includegraphics[width=0.45\linewidth]{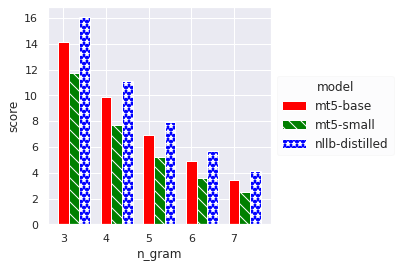}
         \includegraphics[width=0.45\linewidth]{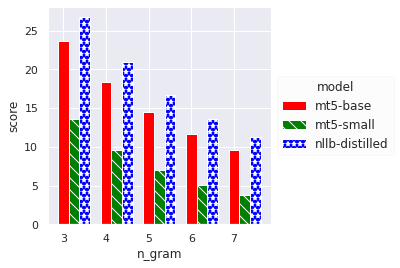}
         \caption{TEP}
         \label{fig:tep-ngram}
     \end{subfigure}

     \begin{subfigure}[b]{1\textwidth}
         \centering
         \includegraphics[width=0.45\linewidth]{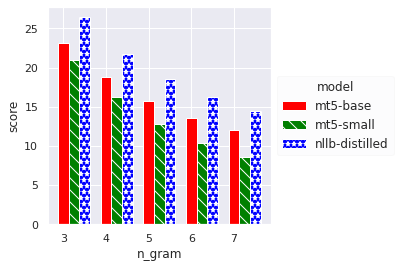}
         \includegraphics[width=0.45\linewidth]{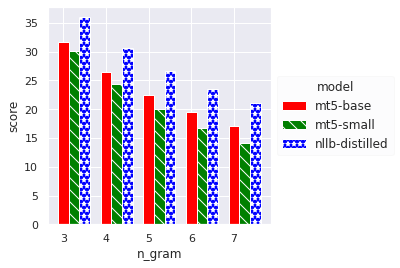}
         \caption{TEP ++}
         \label{fig:quran-ngram}
     \end{subfigure}

     \begin{subfigure}[b]{1\textwidth}
         \centering
         \includegraphics[width=0.45\linewidth]{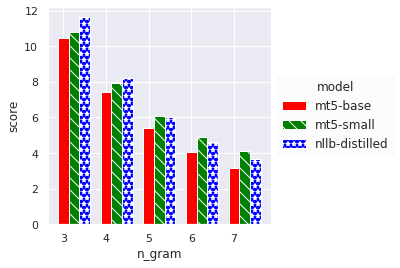}
         \includegraphics[width=0.45\linewidth]{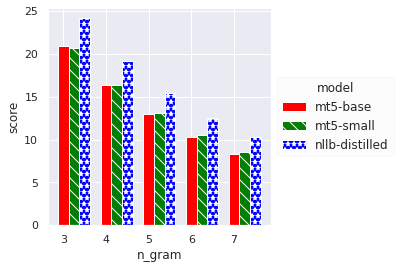}
         \caption{OPUS-100}
         \label{fig:opus-ngram}
     \end{subfigure}
    
    \caption{BLEU Score results for different ngrams separated by translation direction (left side English to Persian and right side Persian to English) and model \textbf{Second part}}
    \label{ngram-results-second}

\end{figure*}
\paragraph{Hardware:} Our Google models were fine-tuned with float32 using TITAN RTX and RTX 3090 Ti GPUs. We used a NVIDIA V100 GPU for the Meta model since it requires a higher level of computation power. The latter was fine-tuned using a PyTorch feature known as automatic mixed precision, which resulted in a reduction in GPU consumption and execution time as opposed to using float16 rather than float32.

\begin{figure}[htp]
    \centering
    \includegraphics[width=1\linewidth]{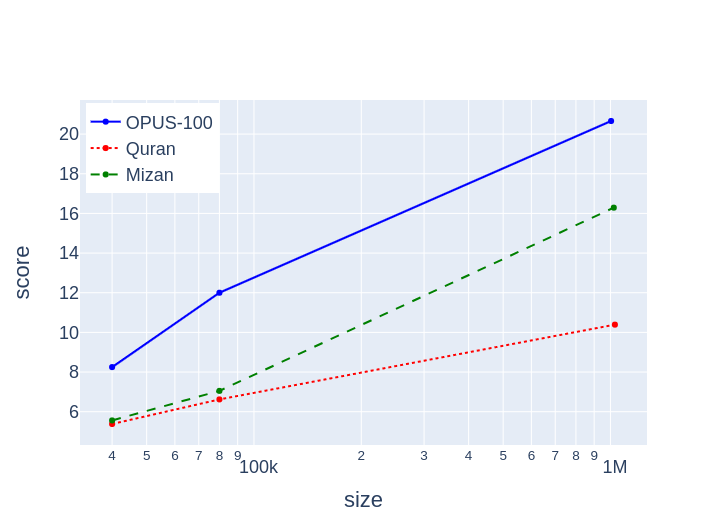}
    \caption{The impact of the number of training instances on the evaluation dataset for translating Persian to English on the mT5 small model.}

    \label{fig:impact-of-training-size}
\end{figure}

\paragraph{Results} Our fine-tuned models were evaluated using SacreBLEU as the evaluation metric. As a result of limited computation power, the maximum sequence length of predicted sentences was smaller than this value for test data. It is not possible to compare real test data with predicted instances with precision. In order to resolve this issue, we truncated test instances that exceeded the maximum sequence length of predicted sentences before calculating the score. Table \ref{final-results} shows the value of SacreBLEU with \(N-gram=3\).

The value of N-grams is an important factor in determining the final BLEU score. This metric utilizes N-grams as contiguous sequences of \{N\} items from a given text sample. To avoid ambiguity and make the results comparable with future research, we report the BLEU measure for \{3, 4, 5, 6, 7\}. Figure \ref{ngram-results-second} illustrates the relationship between N-grams and scores for three models in order to compare their performance and determine the impact of N-grams on their performance. As expected, the results for greater N-grams are lower compared to the smaller ones. In all of the datasets, the Meta NLLB model outperformed both variants of the Google mT5 models. 
\paragraph{Model Evaluation}

\ref{fig:experiments} shows detailed information on experiments about training and validation perplexities, and development BLEU scores during training. 

Training perplexities decreased dramatically from epoch one to two and then followed a gradual decline until epoch seven. However, validation perplexities decreased more rapidly from epoch one to two, and after that, they gradually declined. In some models, this value starts to rise, and models become overfit. Perplexity values in this phase have huge values at the beginning, but they drop after one epoch.

To demonstrate changes in the value of BLEU scores during the training phase and comprehension of the models' performance on each dataset separately, we calculated this value for the development sets per epoch. Most models experience a steady increase, and then tend to decrease or remain flat at this value. However, in three experiments including PEPC bidirectional for mt5-small-fa-en and mt5-base-fa-en, and one directional for mt5-base-en-fa, the evaluation metric dipped at epoch 2 and recovered quickly.

\section{Discussion}
\label{sec:disc}
In this section, some insights into the experiment's outcomes are provided. Additionally, we discuss the quality of the experimented datasets in terms of the number of instances. Figure \ref{fig:impact-of-training-size} shows the maximum BLEU scores for each dataset as a function of the datasets' size in both directions which provides better comparisons of the results.

Generally, datasets like Quran, OPUS-100, and Mizan, with more than one million instances, have received lower or almost the same BLEU score compared to smaller datasets, such as Bible, TEP, TEP++, and PEPC variants.

In comparison to the TEP, the TEP++ dataset achieved a higher score, suggesting that refining noisy instances and increasing the number of instances had a positive impact on the dataset results. In contrast, PEPC dataset variations did not show significant differences between their scores.

Although the Bible is the smallest dataset regarding the total number of instances, it achieved the highest score among all in both translation directions. Another point to be mentioned is that the average sequence length of instances in this dataset is the second largest after the Quran’s average sequence length, but the scores are highly lower for the Quran.

\begin{figure*}
    \centering
    
    \begin{subfigure}[b]{1\textwidth}
         \centering
        \includegraphics[width=0.30\linewidth]{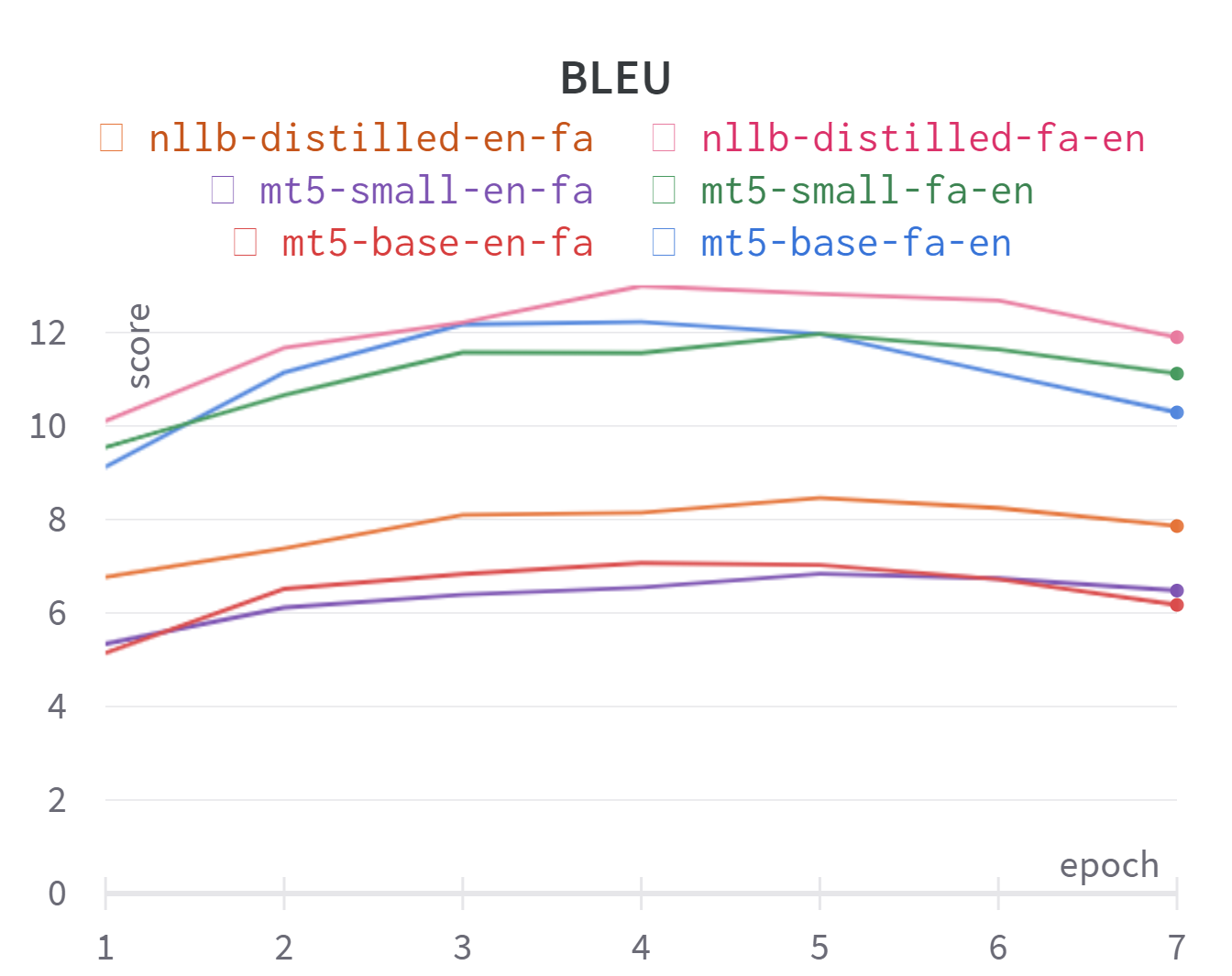}
        \includegraphics[width=0.30\linewidth]{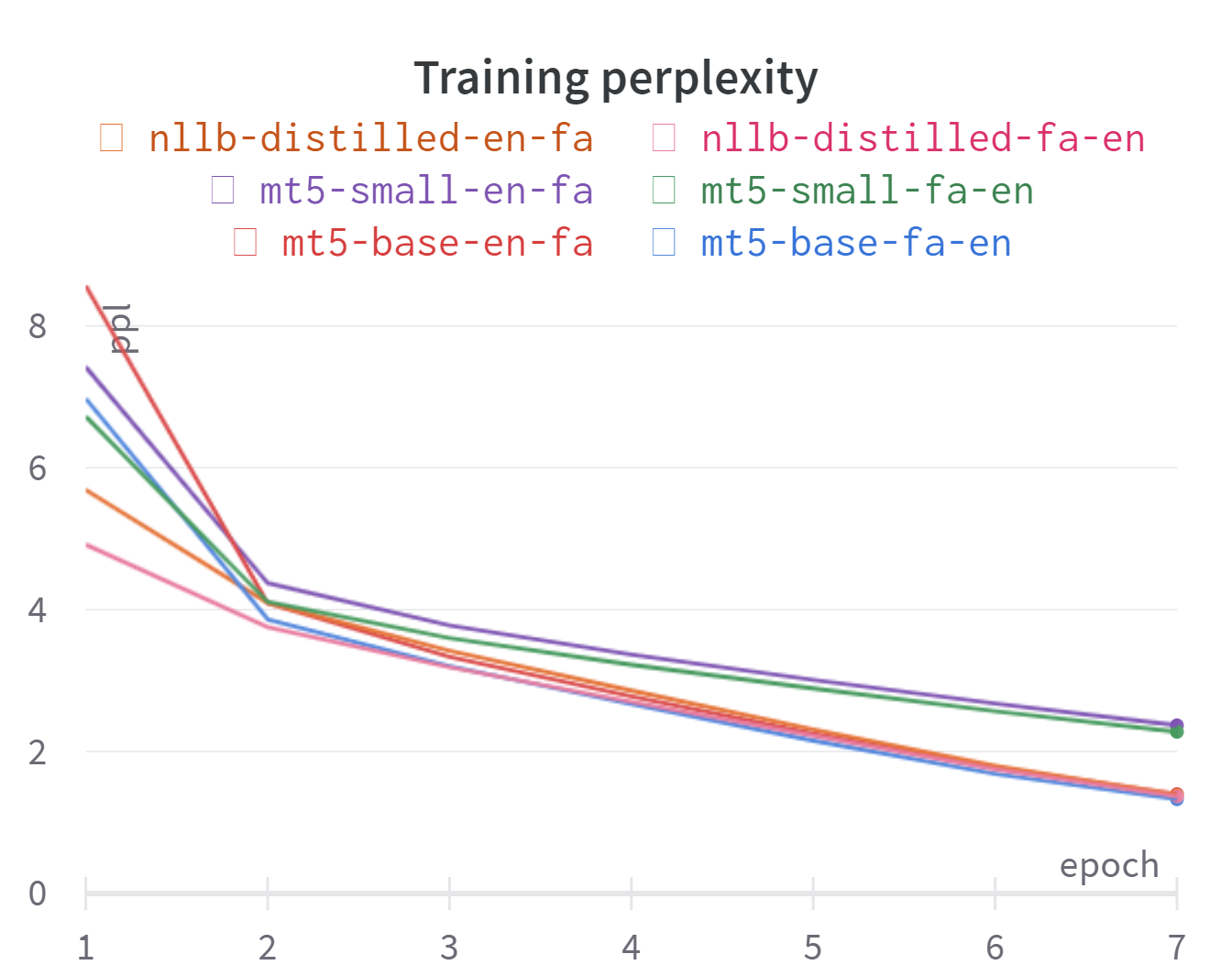}
        \includegraphics[width=0.30\linewidth]{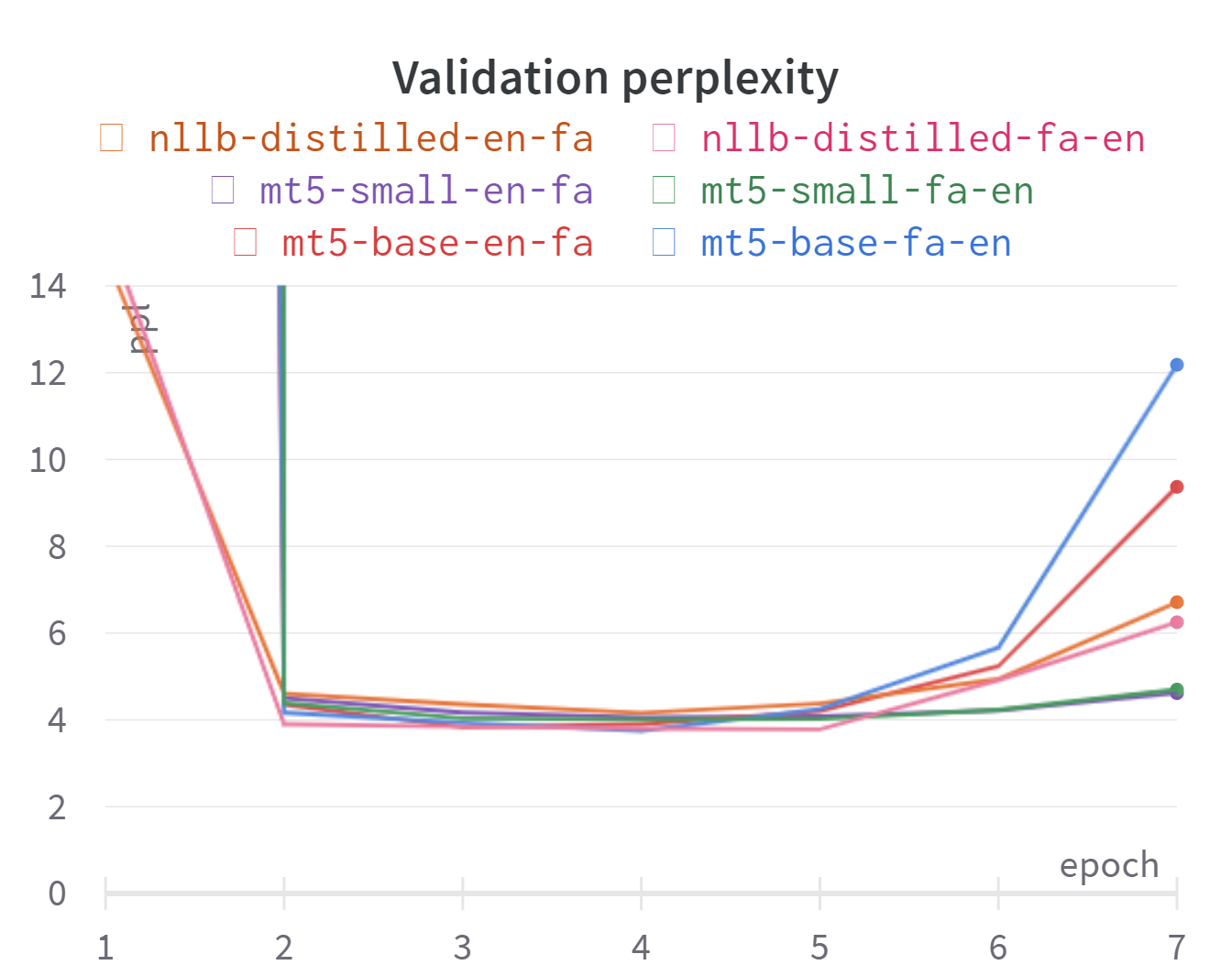}
        \caption{Mizan}
        \label{fig:mizan-exp}
    \end{subfigure}
    \begin{subfigure}[b]{1\textwidth}
         \centering
        \includegraphics[width=0.30\linewidth]{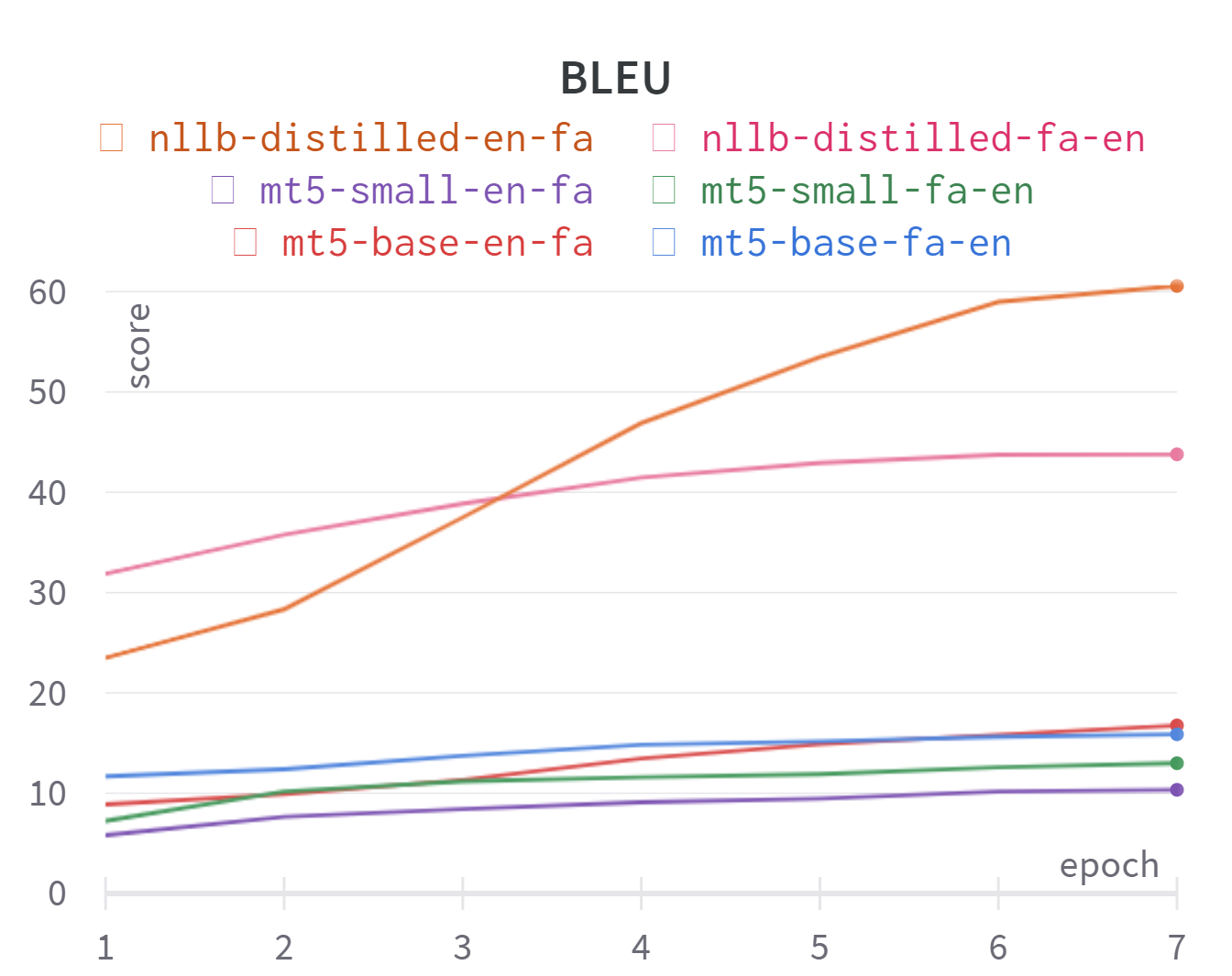}
        \includegraphics[width=0.30\linewidth]{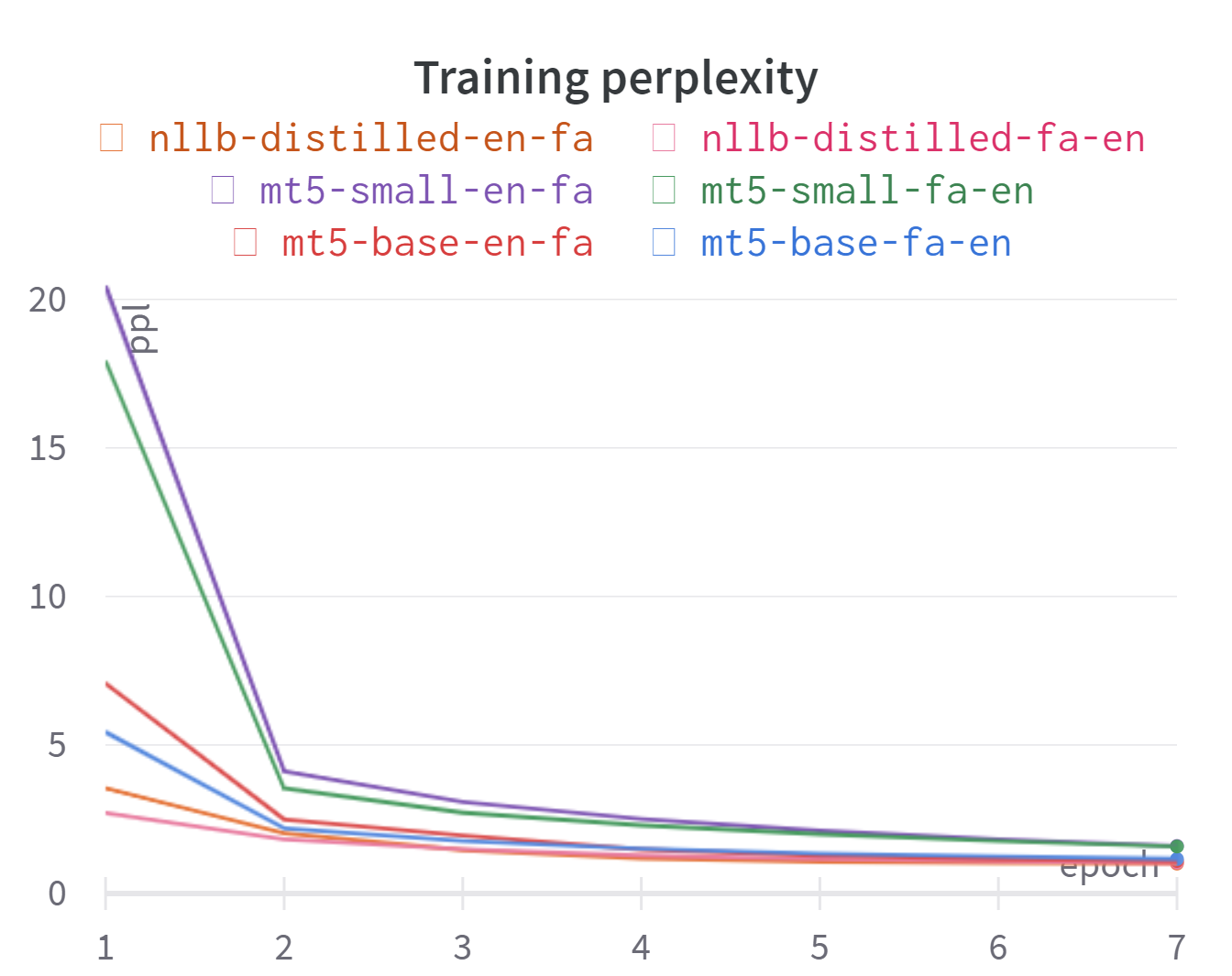}
        \includegraphics[width=0.30\linewidth]{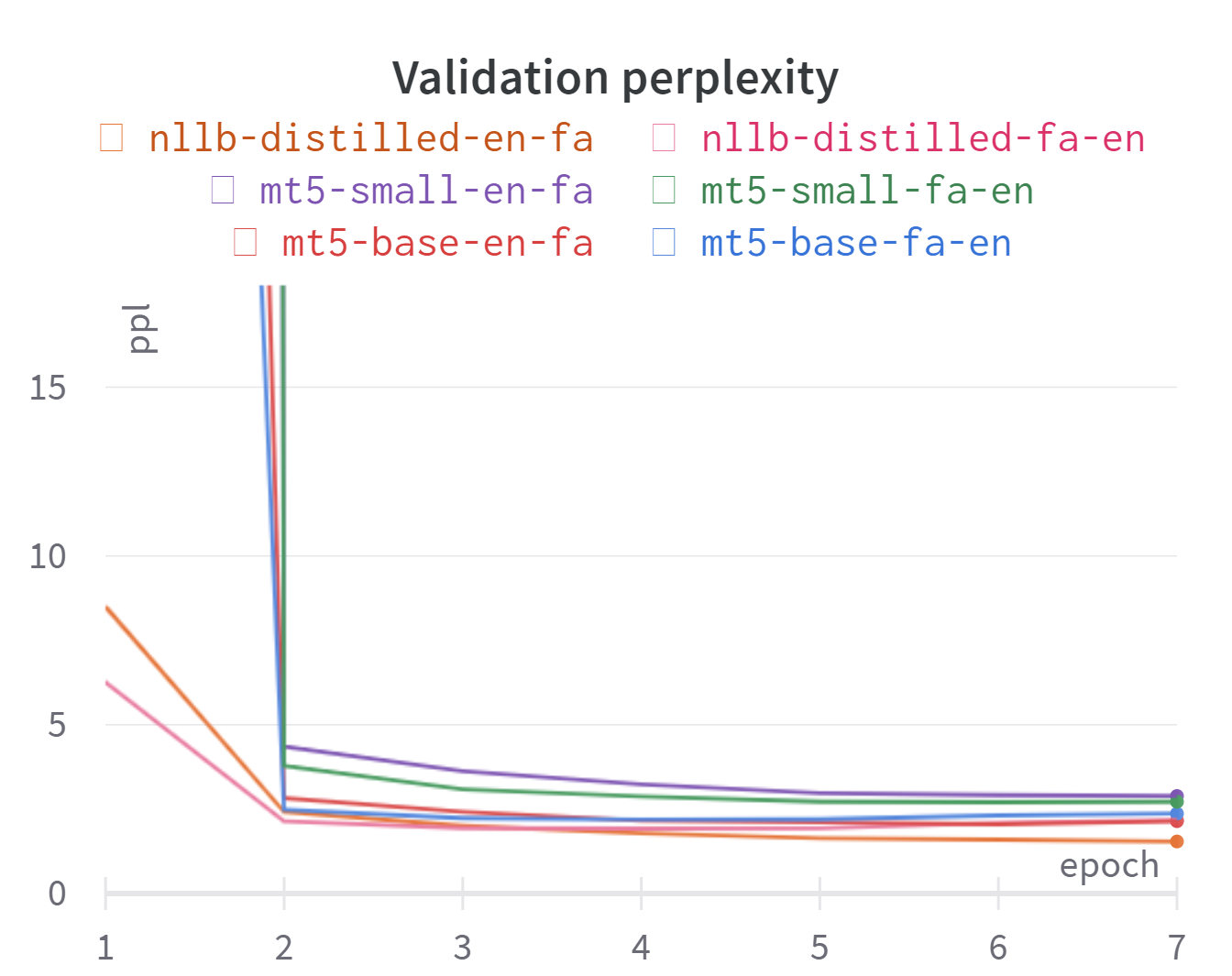}
        \caption{Bible}
        \label{fig:bible-exp}
    \end{subfigure}

    \begin{subfigure}[b]{1\textwidth}
         \centering
        \includegraphics[width=0.30\linewidth]{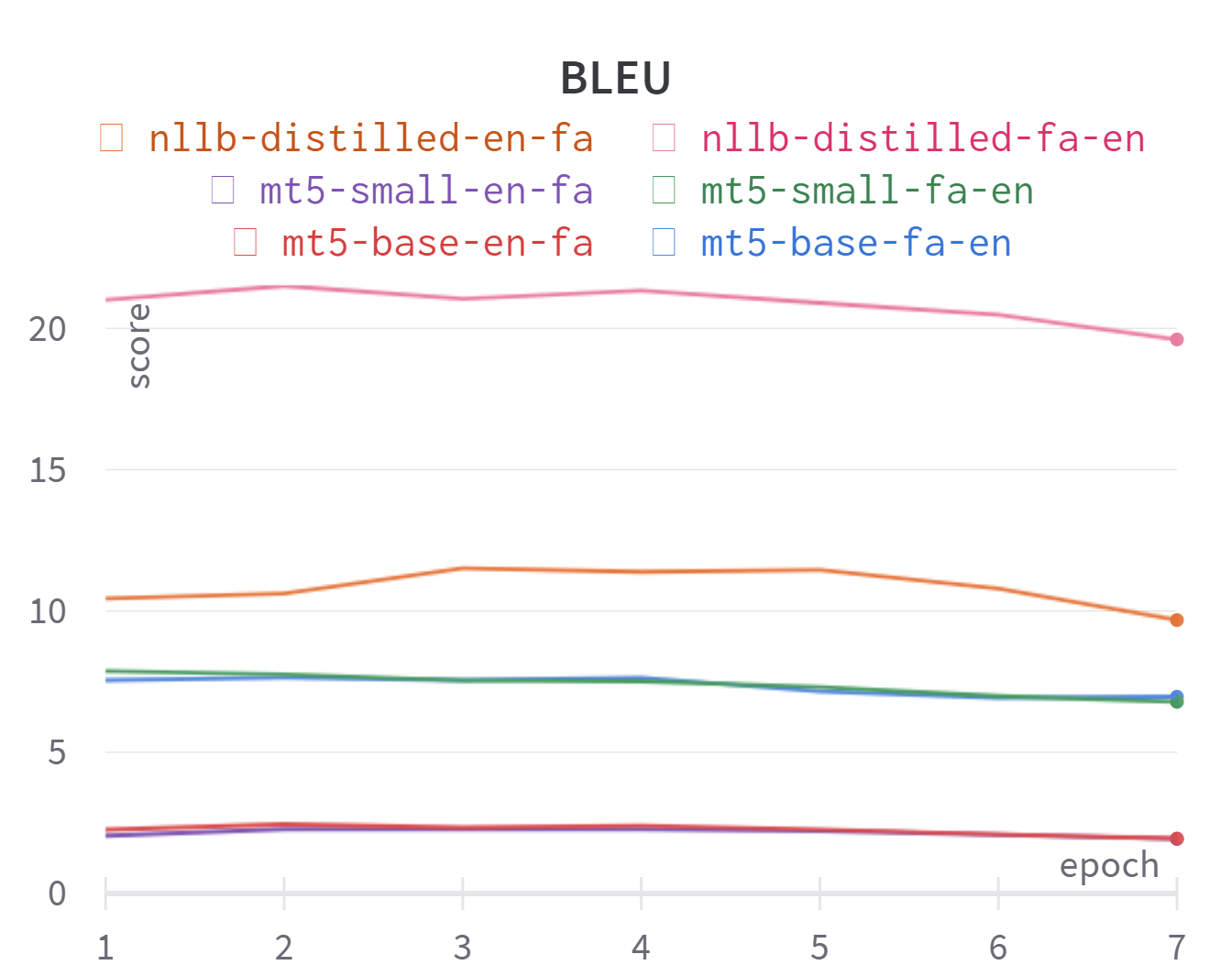}
        \includegraphics[width=0.30\linewidth]{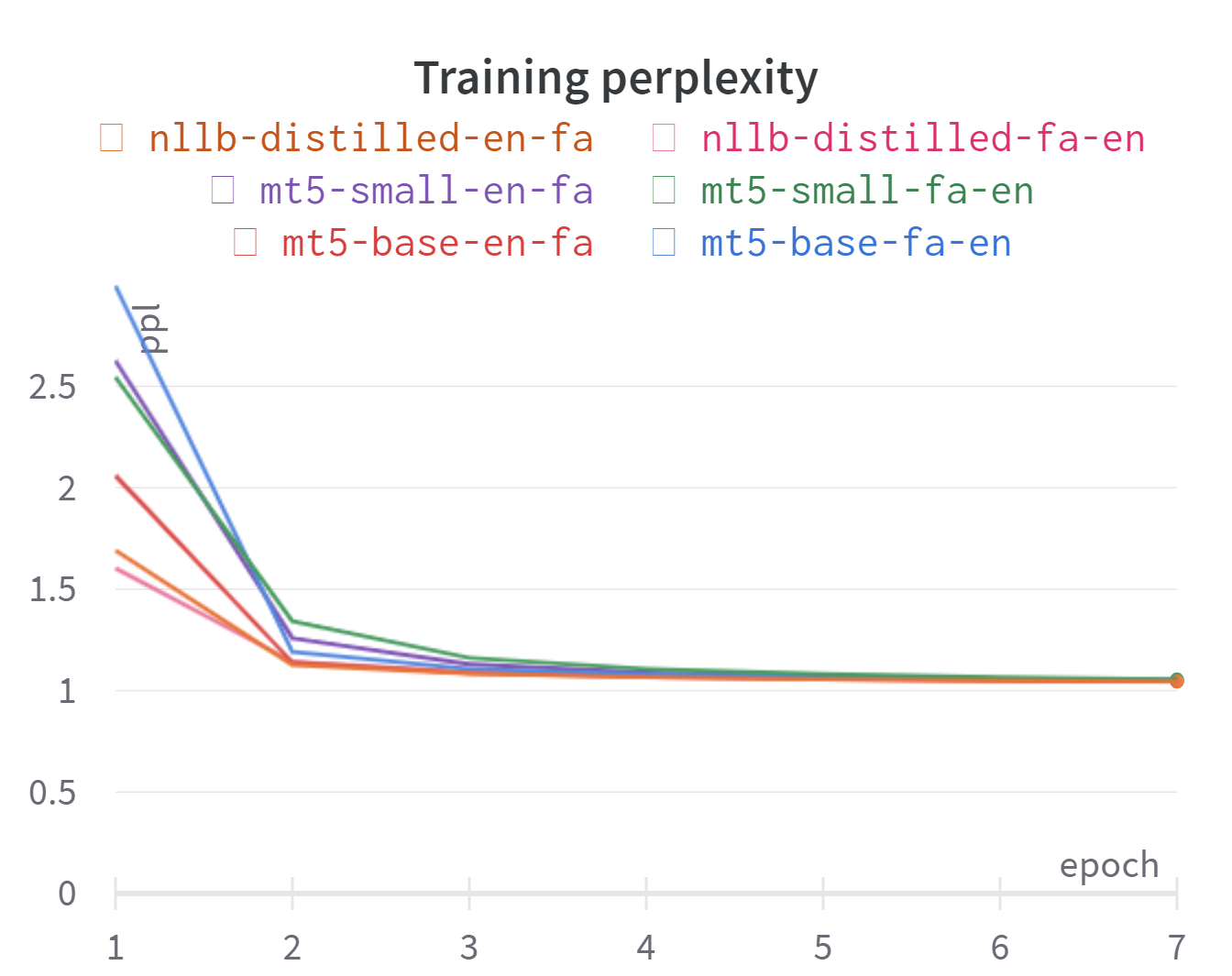}
        \includegraphics[width=0.30\linewidth]{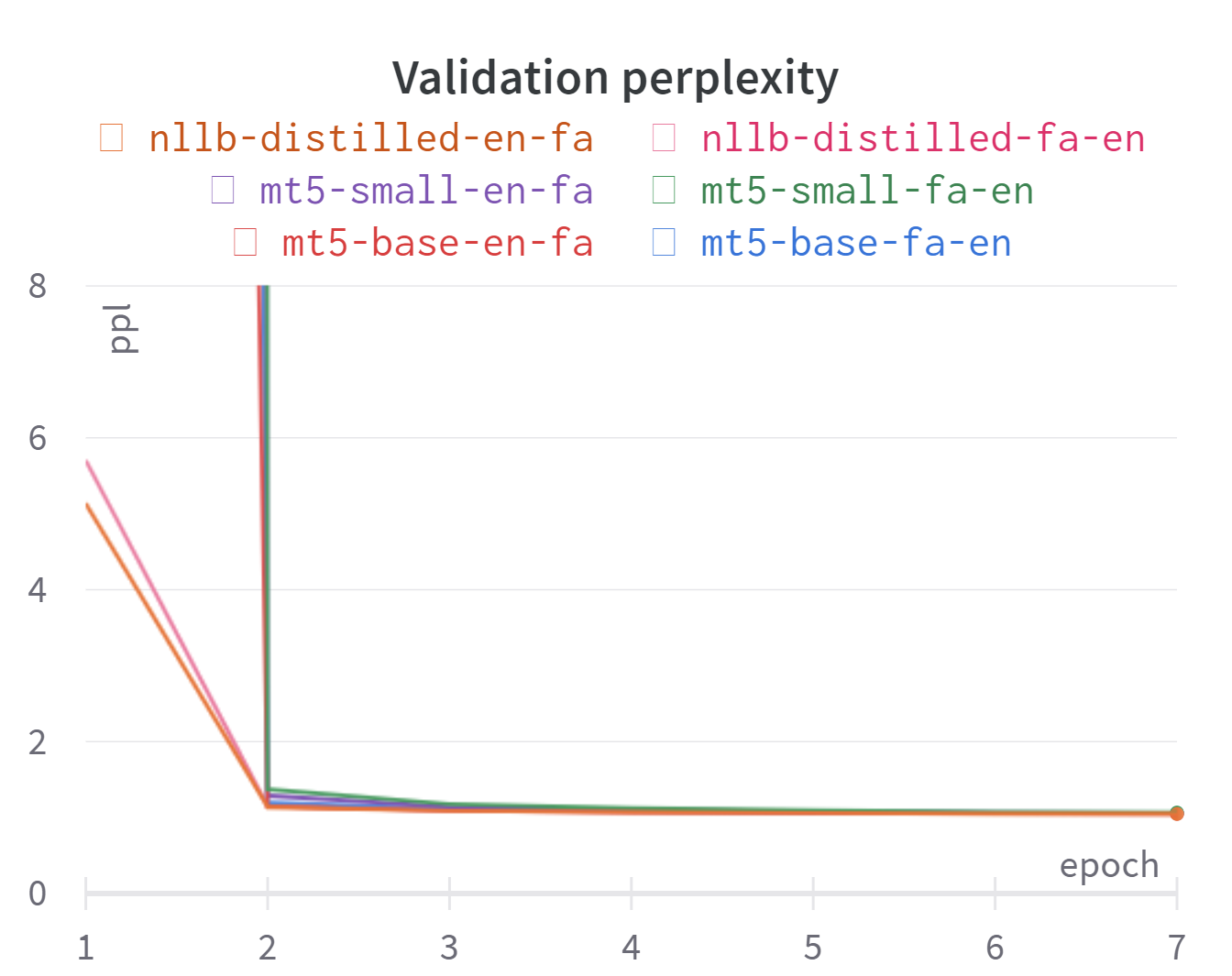}
        \caption{Quran}
        \label{fig:quran-exp}
    \end{subfigure}
    
    \begin{subfigure}[b]{1\textwidth}
         \centering
        \includegraphics[width=0.30\linewidth]{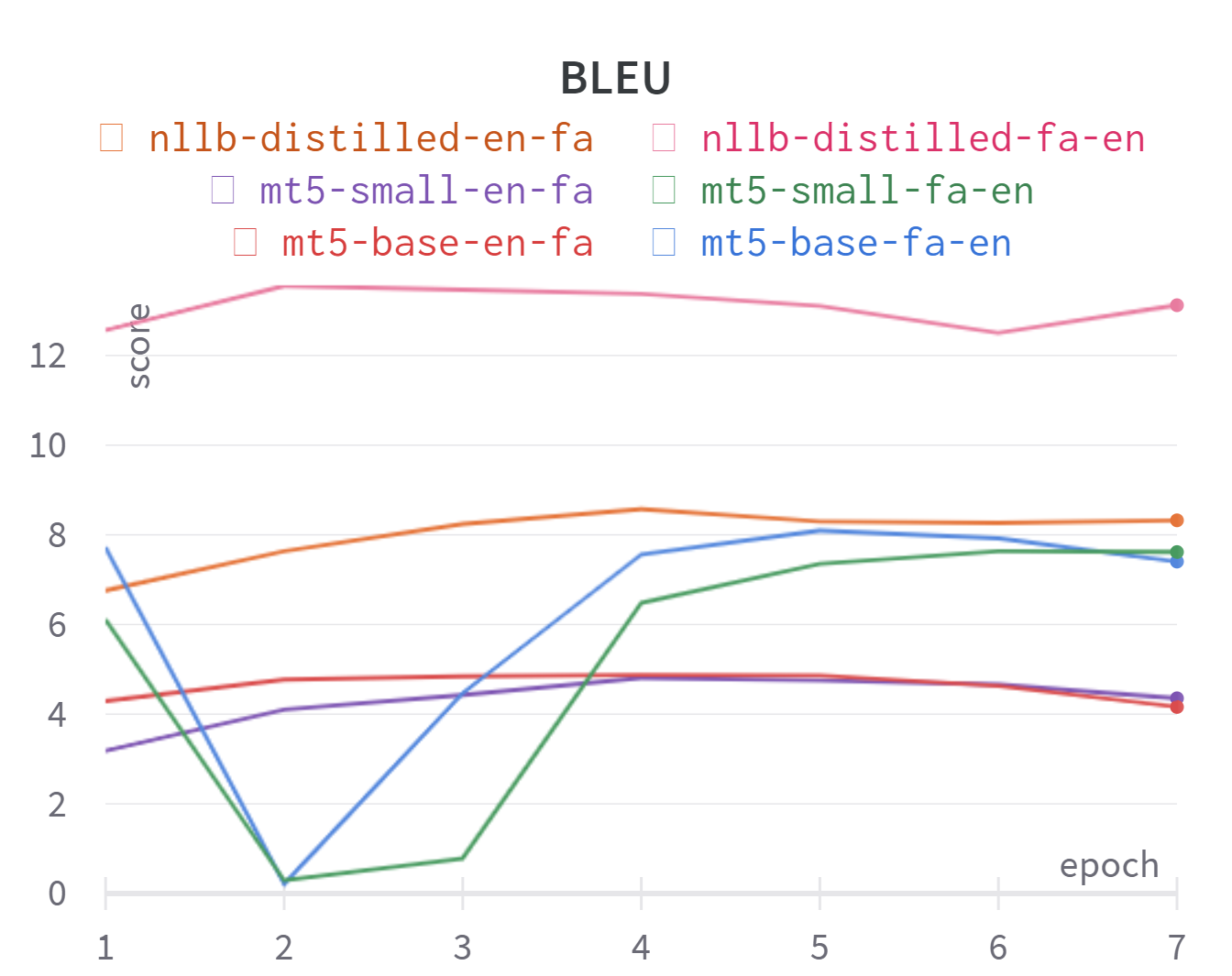}
        \includegraphics[width=0.30\linewidth]{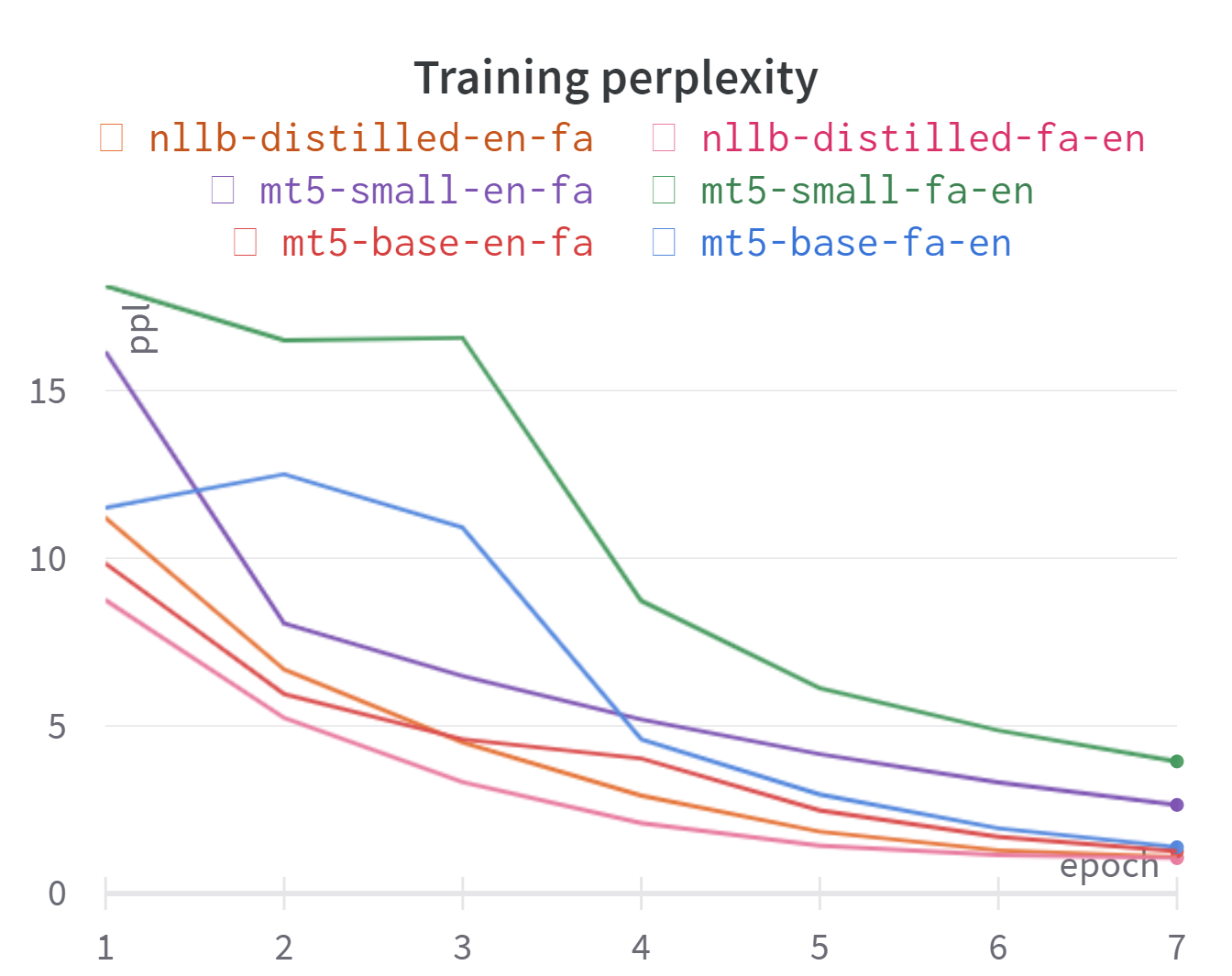}
        \includegraphics[width=0.30\linewidth]{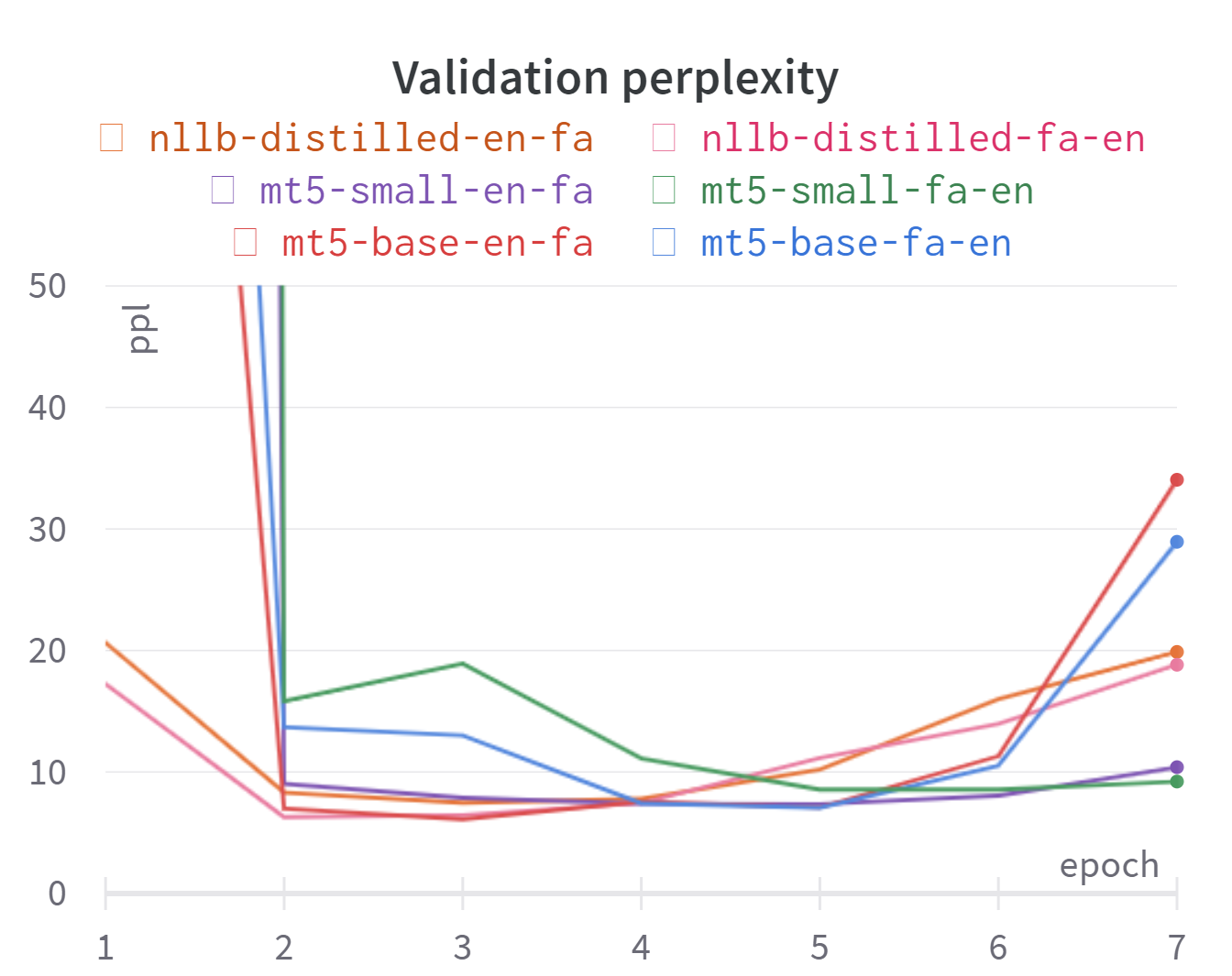}
        \caption{PEPC Bidirectional}
        \label{fig:pepc-bi-exp}
    \end{subfigure}
    \caption{BLEU scores, training perplexities, and validation perplexities for each dataset. \textbf{First part}}
    \label{fig:experiments}
\end{figure*}

\begin{figure*}
    \centering
    \ContinuedFloat
    \begin{subfigure}[b]{1\textwidth}
         \centering
        \includegraphics[width=0.30\linewidth]{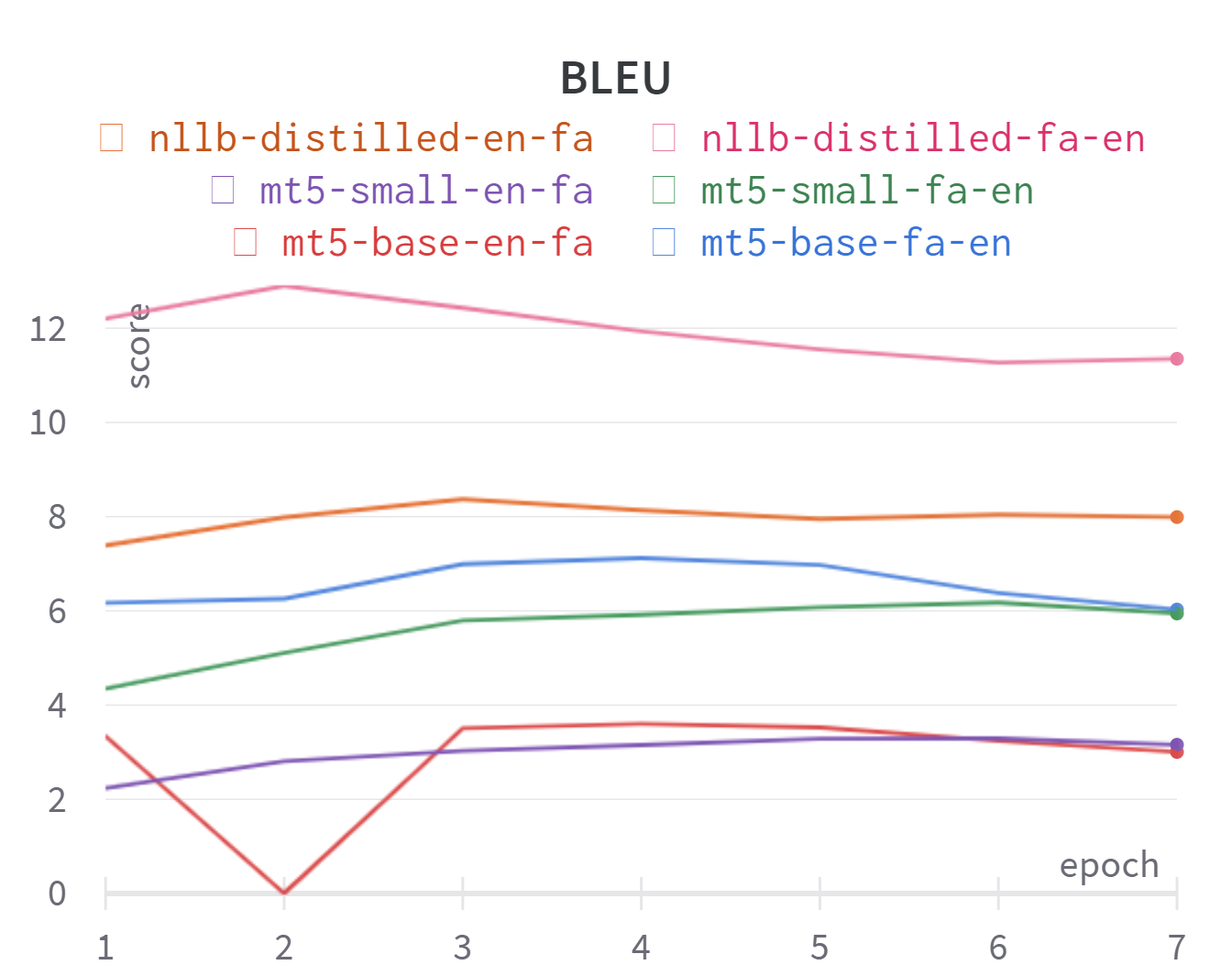}
        \includegraphics[width=0.30\linewidth]{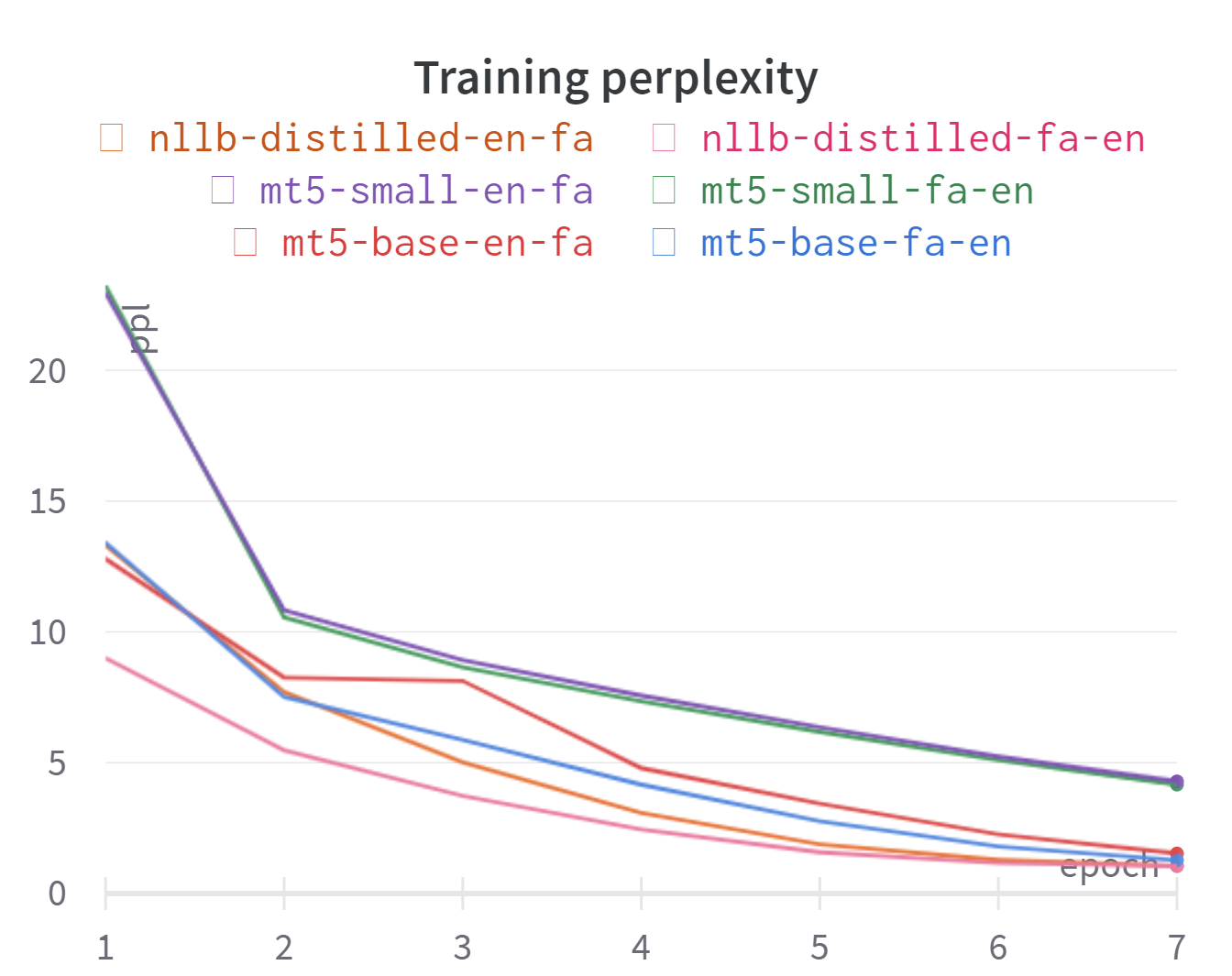}
        \includegraphics[width=0.30\linewidth]{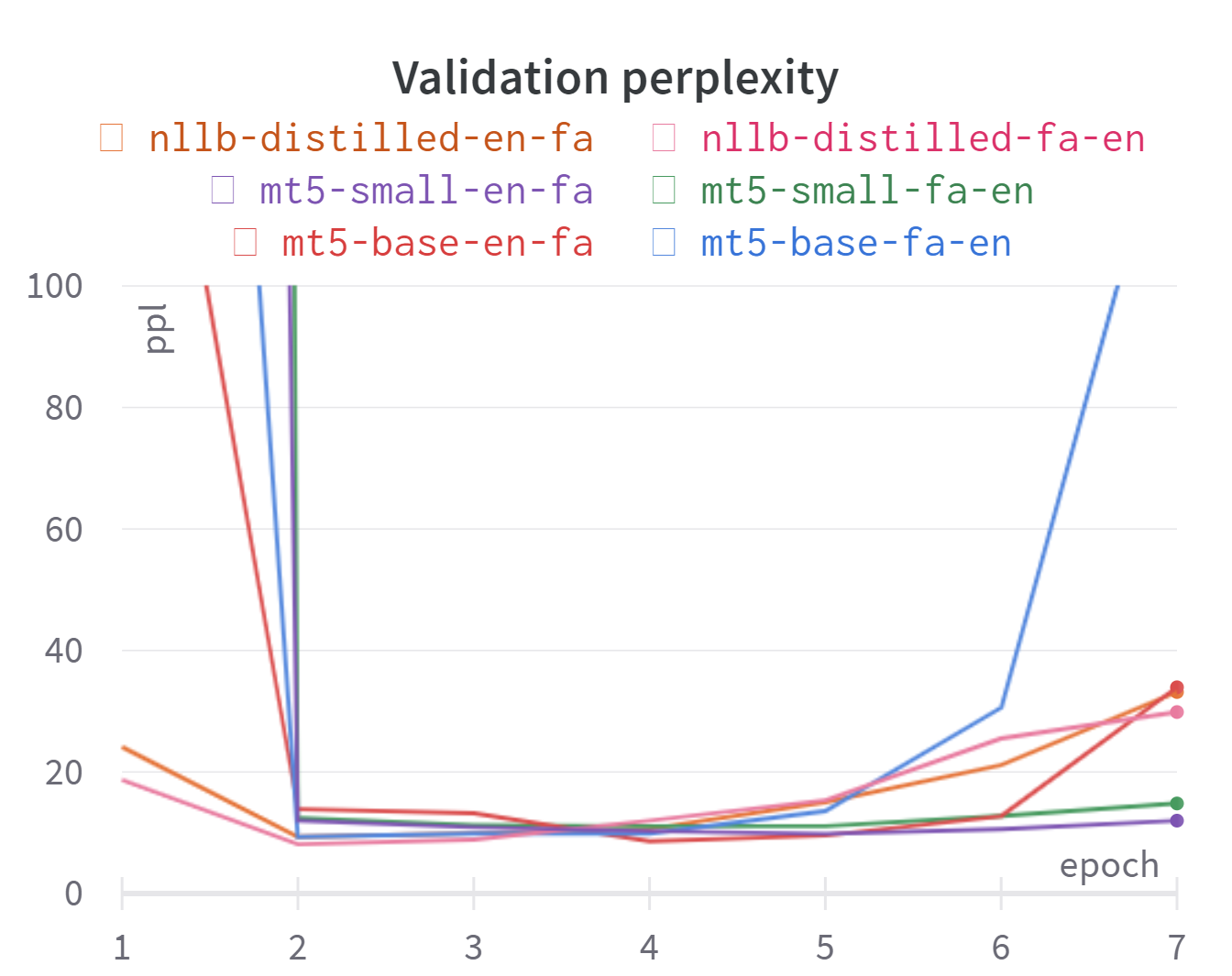}
        \caption{PEPC One Directional }
        \label{fig:pepc-one-exp}
    \end{subfigure}
    \begin{subfigure}[b]{1\textwidth}
         \centering
        \includegraphics[width=0.30\linewidth]{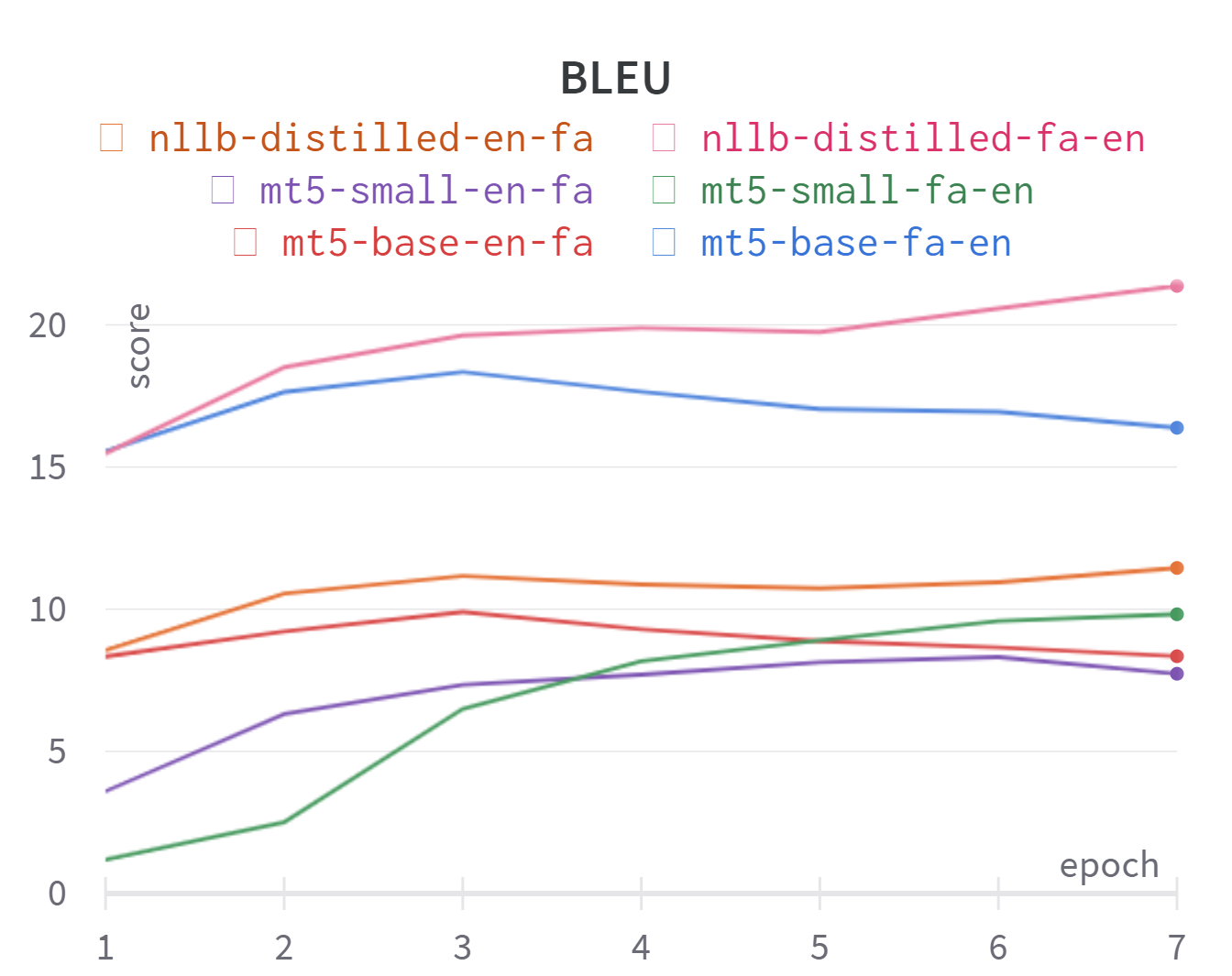}
        \includegraphics[width=0.30\linewidth]{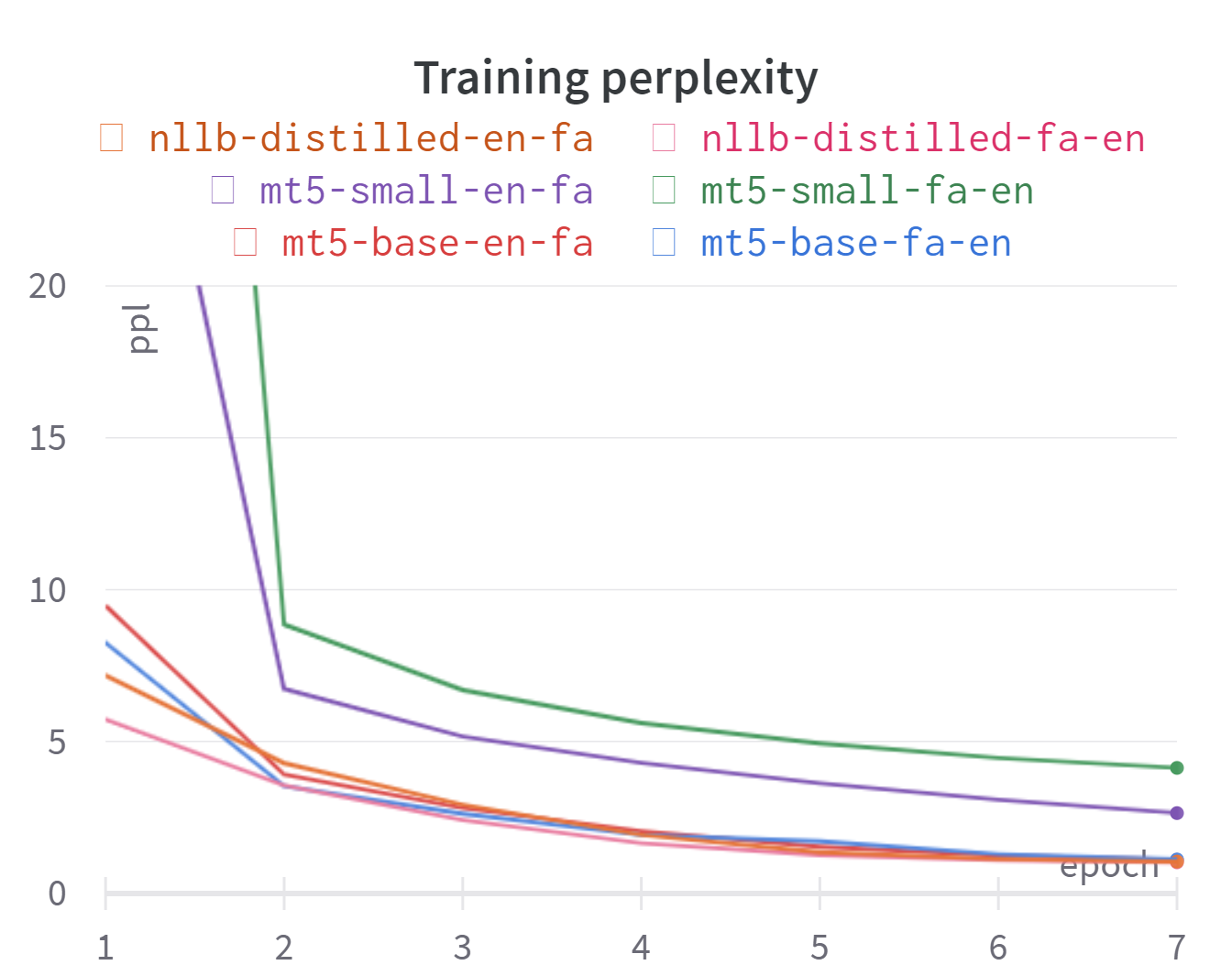}
        \includegraphics[width=0.30\linewidth]{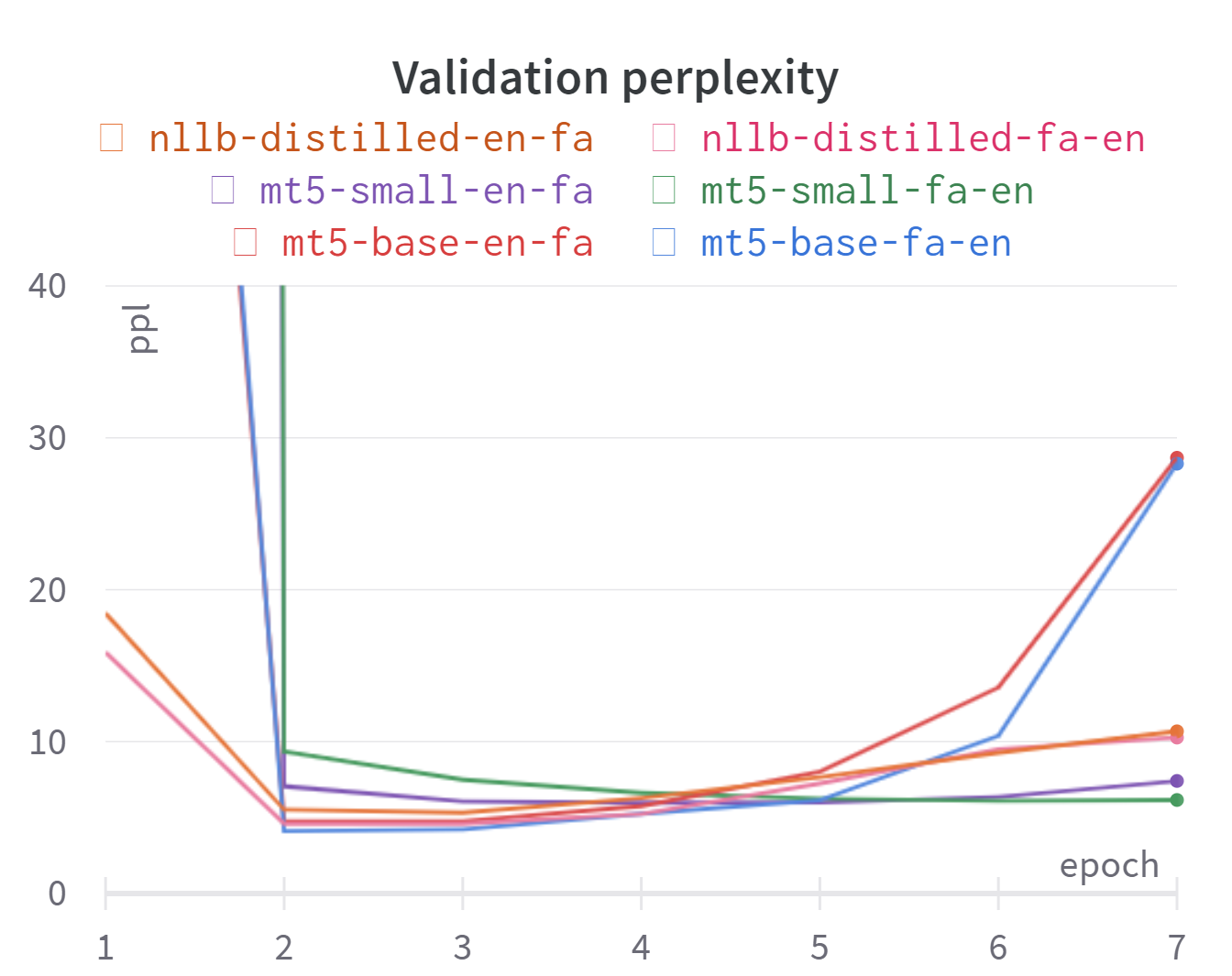}
        \caption{TEP}
        \label{fig:tep-exp}
    \end{subfigure}

    \begin{subfigure}[b]{1\textwidth}
         \centering
        \includegraphics[width=0.30\linewidth]{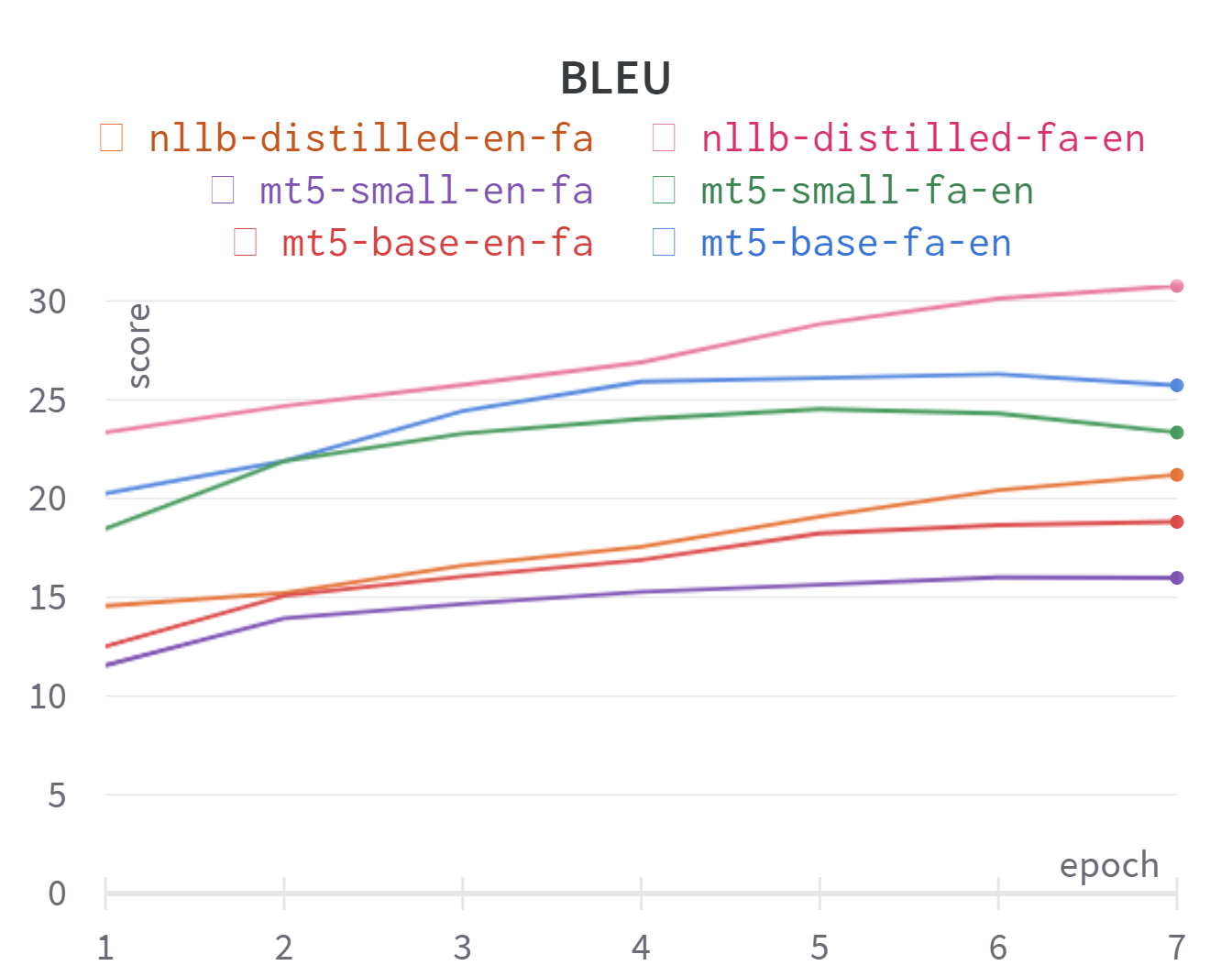}
        \includegraphics[width=0.30\linewidth]{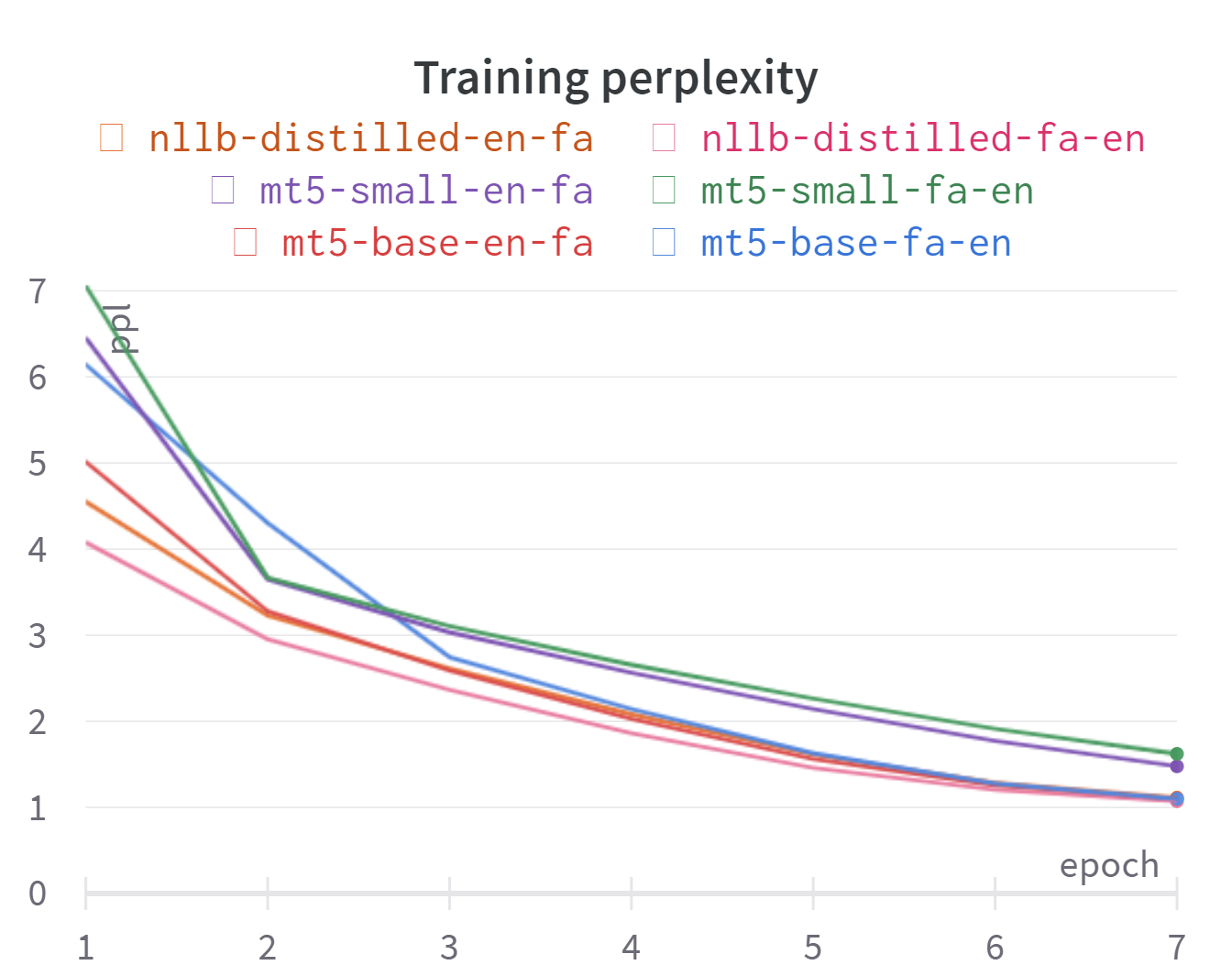}
        \includegraphics[width=0.30\linewidth]{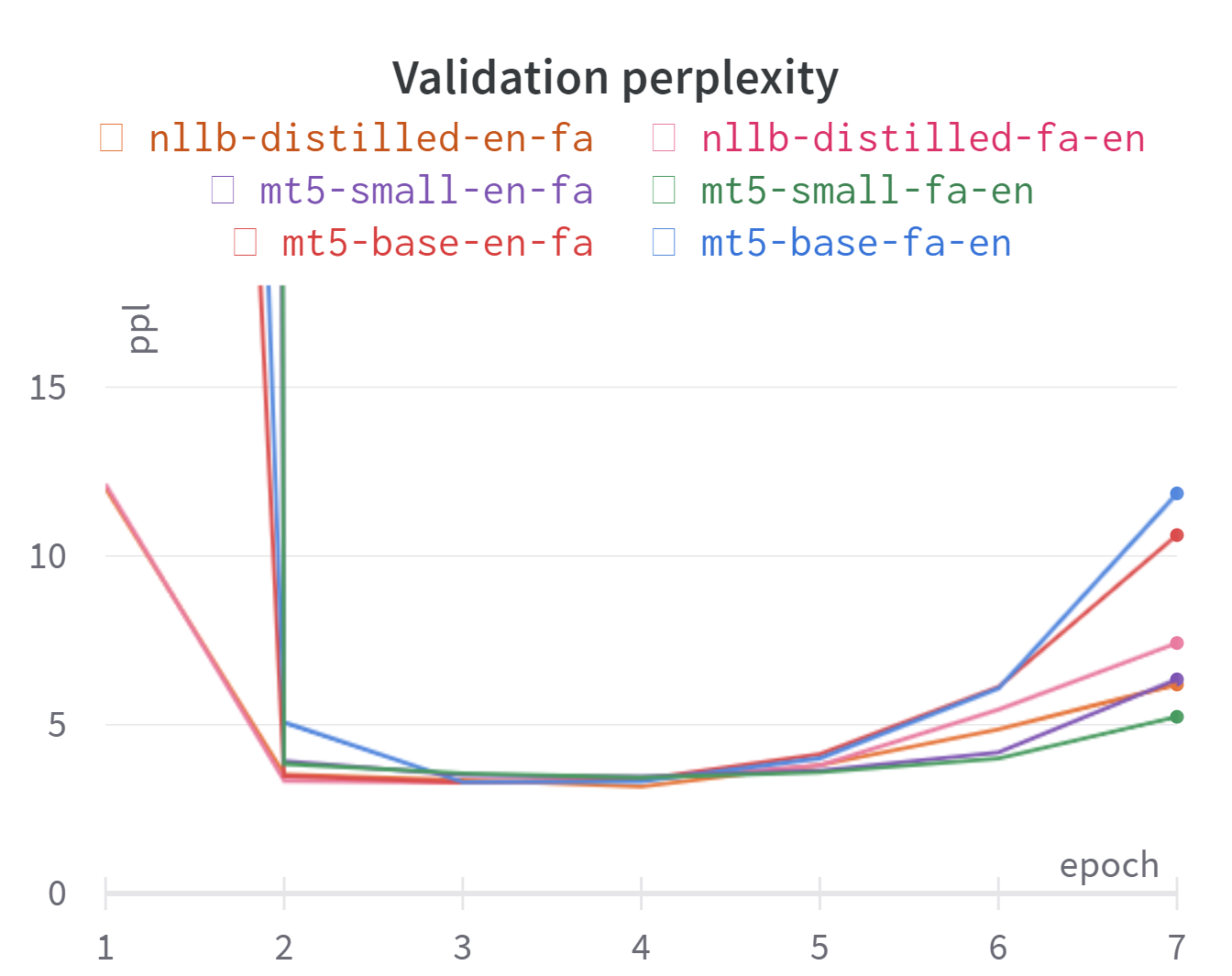}
        \caption{TEP++}
        \label{fig:tep-pp-exp}
    \end{subfigure}
    
    \begin{subfigure}[b]{1\textwidth}
         \centering
        \includegraphics[width=0.30\linewidth]{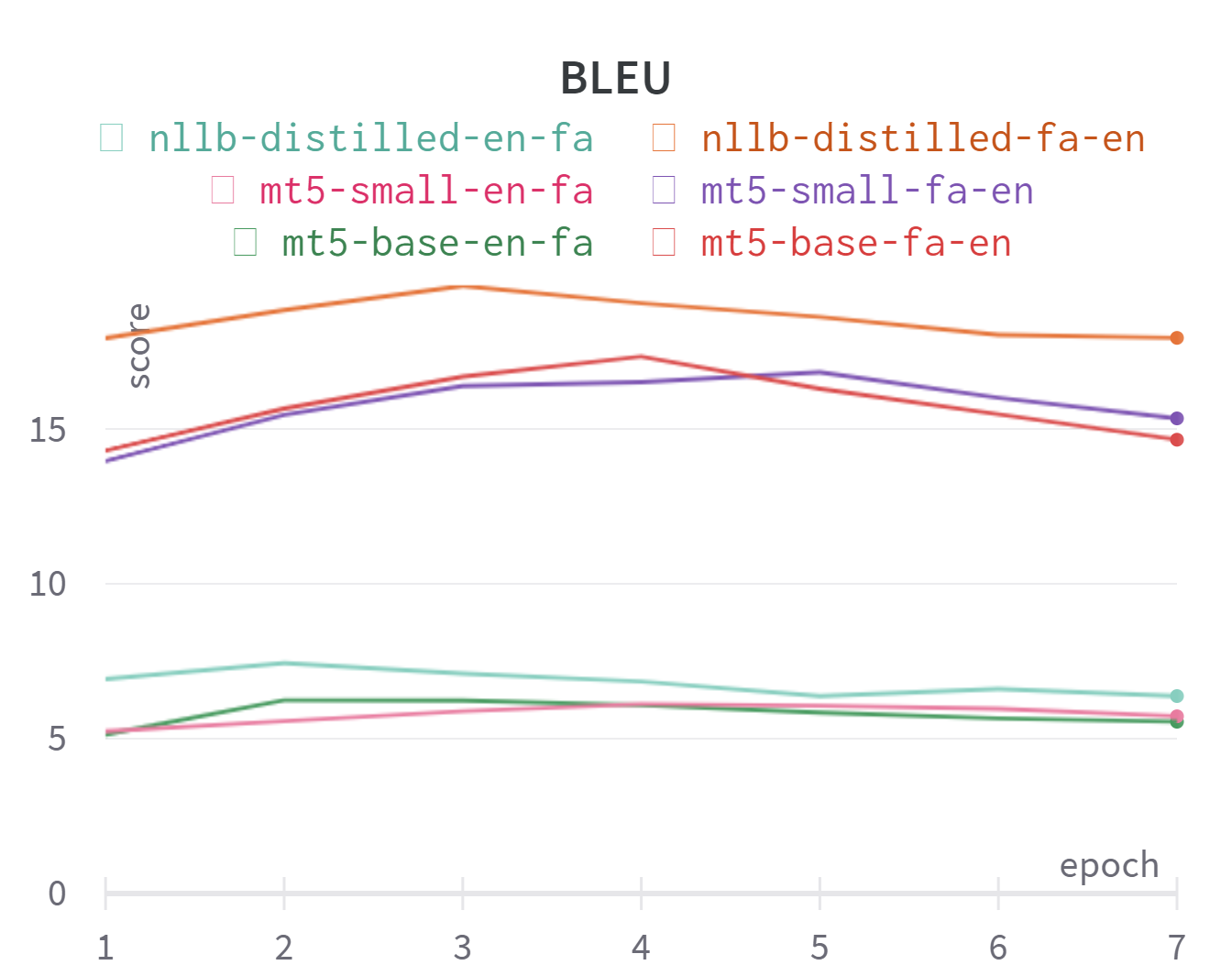}
        \includegraphics[width=0.30\linewidth]{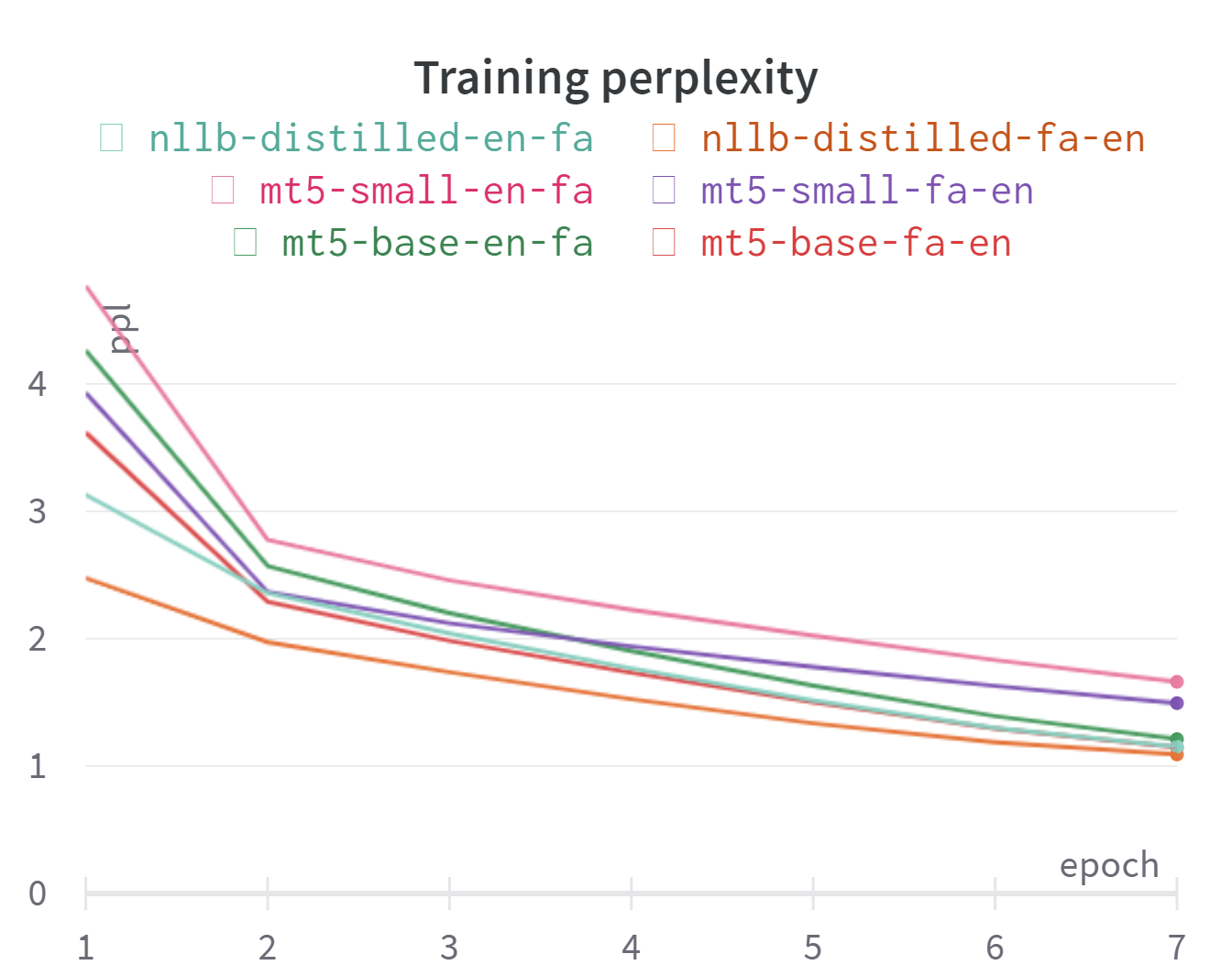}
        \includegraphics[width=0.30\linewidth]{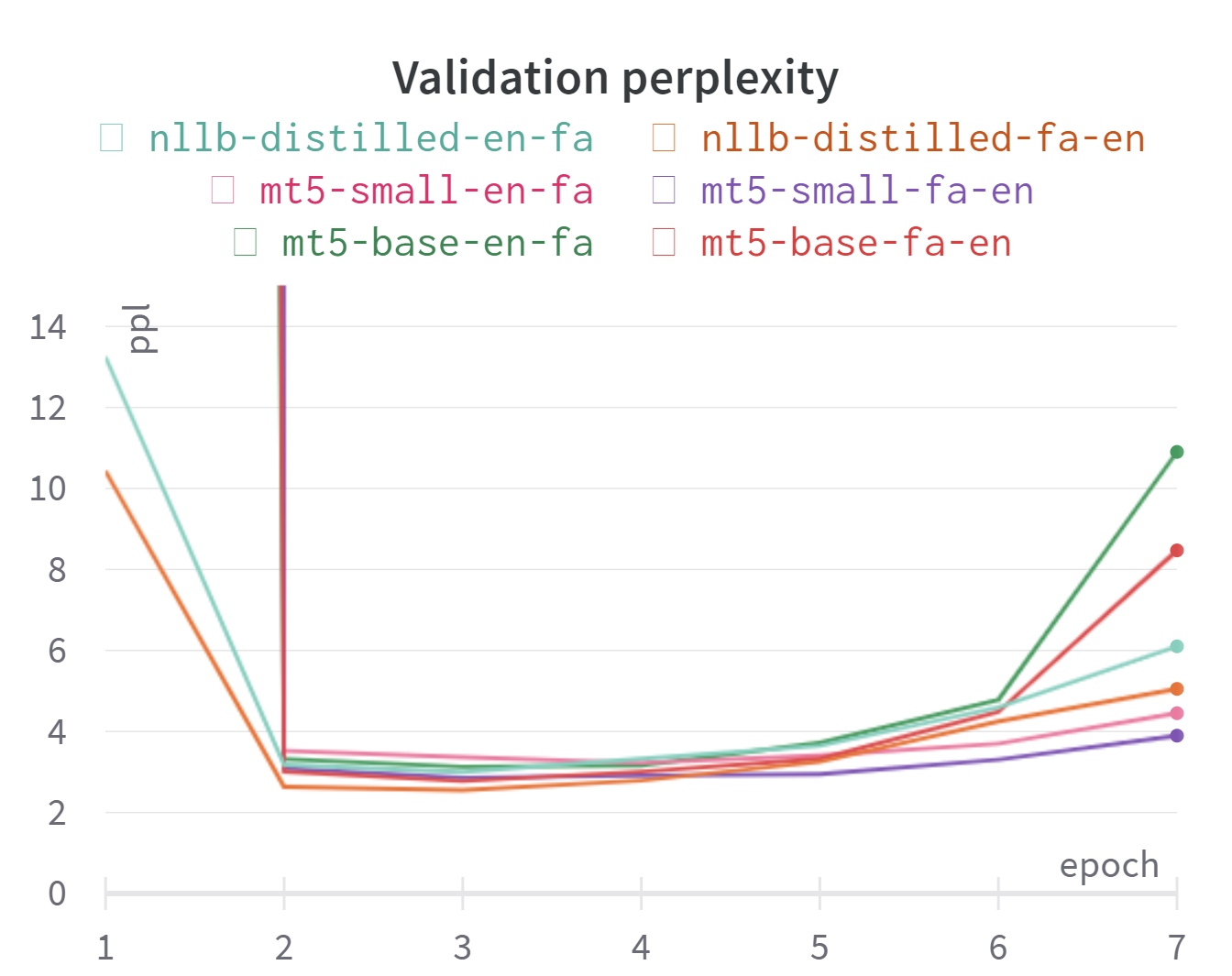}
        \caption{OPUS-100}
        \label{fig:opus-exp}
    \end{subfigure}
    \caption{BLEU scores, training perplexities, and validation perplexities for each dataset. \textbf{Second part}}
    \label{fig:experiments}
\end{figure*}

\paragraph{Quality or Quantity?}

According to the traditional method of improving machine translation results, increasing the size of the training data is expected to increase the value of BLUE score. However, this study indicates that datasets with a higher number of instances tend to achieve lower BLEU scores than datasets with a lower number of instances. Consequently, the quality of the data used for the fine-tuning phase could be more critical than the number of instances. Regarding quality, mistakes in dictation, translations that are not aligned, punctuation errors, and the incorrect word orders in the source and destination directions could change the concept and have a negative effect on the final evaluation value. 

Three datasets with more than one million instances were tested to demonstrate how the number of training samples affects the value evaluation metric. From those datasets, we sampled 40k and 80k instances and fine-tuned the Google mT5 small model. Based on this experiment, figure \ref{fig:impact-of-training-size} shows that by increasing the number of instances, the model shows better results.

\paragraph{Translation Direction}
The general order of obtained BLEU scores in both directions is almost identical. There are a few factors we should take into account. The Bible dataset represented the highest BLEU score in both directions. However, in the English-to-Persian direction, the OPUS-100 dataset had the lowest BLEU score, and the one-directional PEPC dataset had the lowest BLEU score in Persian-to-English direction. 
Although almost all datasets performed better in Persian-to-English translation, the Bible dataset performed significantly better in English-to-Persian translation by near 20\% higher BLEU score. In the Persian-to-English translation, the OPUS-100 dataset performs significantly better than the Mizan dataset, while in the opposite direction, the Mizan dataset shows greater performance.

\section{Conclusion}
\label{sec:conc}
In this study, we reviewed a majority of Persian-English parallel corpora and established standard baselines for eight datasets. The datasets are evaluated using two multilingual seq2seq models based on a transformer architecture. Our analysis of 48 experiments indicates that the Bible and PEPC datasets have the highest and lowest BLEU scores, respectively. Additionally, we conclude that Meta's basic variant outperforms previous transformer-based approaches by a significant margin. The findings also indicate that in most experiments, the evaluation metric for translation from Persian to English is higher than the evaluation metric for translation from English to Persian.  
To the best of our knowledge, this is the first study that represents baselines for each dataset separately by seq2seq models. We hope that this research will assist researchers to compare their methods with the baselines and evaluate them specifically for the Persian language.

\label{sec:bibtex}
\section*{Acknowledgements}
This work has been supported by the Simorgh Supercomputer - Amirkabir University of Technology under Contract No ISI-DCE-DOD-Cloud-900808-1700.
\bibliography{anthology,acl_latex}
\bibliographystyle{acl_natbib}





\end{document}